\documentclass[11pt]{article}

\usepackage[preprint]{acl}

\usepackage{times}
\usepackage{latexsym}

\usepackage[T1]{fontenc}

\usepackage[utf8]{inputenc}

\usepackage{microtype}

\usepackage{inconsolata}

\usepackage{graphicx}

\usepackage{amsthm}
\usepackage{amssymb}
\usepackage{amsmath}
\theoremstyle{definition}
\newtheorem{definition}{Definition}[section]
\usepackage{booktabs}
\usepackage{multirow}
\usepackage{xcolor}

\definecolor{myblue}{HTML}{4682B4}
\title{\textsc{SwiftLoop}: Selective Refinement for Adaptive Implicit Reasoning in Looped Transformers}

\author{
 \textbf{Tianyi Zhao\textsuperscript{1}},
 \textbf{Ke Wan\textsuperscript{1}},
 \textbf{Chen Chen\textsuperscript{1}}
\\
 \textsuperscript{1}University of Virginia
\\
\texttt{\{abs4dj,tbn5pj,zrh6du\}@virginia.edu}
}

\begin{document}
\maketitle
\begin{abstract}
Looped Transformers advance implicit chain-of-thought reasoning through recurrent latent refinement.
However, existing looped reasoners typically employ fixed-depth recurrence, updating all latent positions at every iteration and incurring redundancy when further refinement is unnecessary.
We introduce \textsc{SwiftLoop}, an inference-time framework for selective recurrent refinement through position-wise freezing, enabling adaptive computation
allocation across both recurrent
depth and active width.
SwiftLoop learns lightweight freezing policies through hindsight-guided supervision derived from two complementary criteria: K/V reusability at the internal representation level and functional impact at the output level.
At inference time, these policies guide adaptive freezing based on observed hidden states.
Extensive experiments demonstrate that
SwiftLoop achieves favorable
accuracy--computation trade-offs
over fixed-loop inference and
adaptive baselines,
with measured inference acceleration.
Beyond implicit reasoning, we further demonstrate its applicability to recurrent prompt prefilling in general looped models.
\end{abstract}

\section{Introduction}
\begin{figure*}[t]
\centering
  \includegraphics[width=0.99\textwidth]{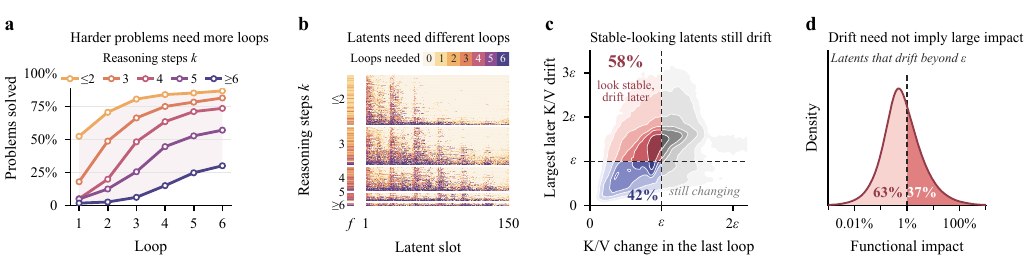}
  \caption{\textbf{Analysis of recurrent computation dynamics.} (a) The loop depth sufficient for final predictions varies with difficulty.
  (b) The loop depth required by individual latent slots varies.
  (c) locally stable K/V may exhibit substantial future drift. (d) Substantial K/V drift need not correspond to large effects on the prediction margin.}
  \label{fig:motiv}
\end{figure*}
Implicit chain-of-thought (CoT) reasoning has emerged as an efficient alternative to explicit CoT, replacing token-by-token reasoning traces with continuous intermediate states~\cite{wei2022chain,deng2023implicit,deng2024explicit}. 
Representative approaches include Coconut~\citep{hao2025training}, which employs curriculum learning to progressively replace explicit reasoning steps with latent states, and CODI~\citep{shen-etal-2025-codi}, which learns a latent reasoning pathway through self-distillation from an explicit CoT pathway.
Despite substantial effort, implicit reasoning methods have yet to match the performance of explicit CoT on challenging reasoning tasks~\cite{he2025semcot,wei2026simcot}.
%

One direction is to increase internal computation. Looped Transformers~\cite{dehghani2018universal,geiping2025scaling,zhu2025scaling,saunshi2025reasoning} thus offer a natural mechanism by repeatedly applying shared modules for recurrent internal state refinement, increasing effective computational depth while reusing model parameters. 
Building on this framing, \citet{fan2026bridging} introduce LOTUS, which iteratively refines a parallel workspace of latent slots, using token-level supervision from explicit CoT steps during training, and close much of the performance gap between implicit and explicit CoT reasoning.
Despite effectiveness, it employs a fixed inference strategy: every instance executes the same number of loops, and every latent slot is updated at each iteration, regardless of whether further refinement is necessary, incurring redundant computation (Figure~\ref{fig:motiv}a,b).

This motivates adaptive computation allocation~\cite{bae2025mixtureofrecursions,wang2026recurtrace,yang2026flashloop,fu2026thinkathard}, which remains challenging along both {depth} and {width}: neither the number of loops sufficient for each instance nor the latent slots that still benefit from further updates are known in advance at inference time.
Position-wise freezing offers a unified strategy for adapting both dimensions: a latent slot can cease further updates while its cached keys and values (K/V) remain accessible to active positions through attention.
%
The central challenge is thus determining when further refinements can be safely omitted.
Addressing this challenge intuitively entails two perspectives: \textbf{internal representation level} and \textbf{output function level}.
For the former, one straightforward way is to consider K/V reusability. Inter-iteration state variation provides a natural but retrospective convergence signal~\cite{yang2026flashloop}. However, stability observed thus far may precede future K/V drift (Figure~\ref{fig:motiv}c). 
For the latter, functional impact can be characterized by how a freezing decision affects the output, which also involves relating the current state to the future prediction outcome.
Importantly, the two perspectives are distinct, as K/V drift need not imply a large functional impact on the prediction (Figure~\ref{fig:motiv}d).
Together, they motivate hindsight-guided freezing decisions for adaptive looping that spans both levels.
%

To this end, we propose \emph{SwiftLoop}, an adaptive looping policy that enables position-wise selective freezing for adaptive computation at inference.
After each loop, a lightweight predictor determines which active slots to freeze based on their observed hidden states, thereby regulating per-position effective depth and per-loop active width. 
We investigate two alternative supervision objectives for learning this freezing policy:
{K/V reusability}, which characterizes the future consistency of cached K/V; and {functional impact}, which estimates the effect of freezing a latent slot on the model's final prediction.
Beyond implicit reasoning, we further extend this selective freezing policy to recurrent prompt prefilling, 
broadening its scope from specialized latent slots to general context-token states.
We evaluate SwiftLoop across four benchmarks using LOTUS~\cite{fan2026bridging}, a looped model for implicit reasoning, at varying scales, and further examine its applicability to recurrent prompt prefilling on two general looped models, Ouro~\cite{zhu2025scaling} and Huginn~\cite{geiping2025scaling}, across three datasets.
Extensive experiments demonstrate that SwiftLoop improves computational adaptivity while maintaining competitive task performance relative to fixed-loop inference.

\section{Problem Formulation}
\label{sec:formulation}

\paragraph{Looped Models.}
We consider a looped Transformer with $L$ layers operating over $N$ positions, represented by $\mathbf X=(x_1,\ldots,x_N)$.
The model performs an initial pass ($t=0$) followed by $R$ recurrent iterations:
\begin{equation}
\begin{aligned}
    \mathbf H^{(0)}&=\mathcal I(\mathbf X;\mathbf c),\\
\mathbf H^{(t)}&=F_\theta(\mathbf H^{(t-1)};\mathbf X,\mathbf c),\, t=1,\ldots,R,
\end{aligned}
\end{equation}
where $\mathcal I$ denotes the model-specific initialization operator, $F_\theta$ is the shared Transformer block and $\mathbf c$ denotes optional fixed conditioning information.
Let $h_i^{(t)}$ denote the hidden state of position $i$ after pass $t$, and $\kappa_{i,\ell}^{(t)}=[K_{i,\ell}^{(t)};V_{i,\ell}^{(t)}]$ its concatenated key-value (K/V) vectors at layer $\ell$.

In \textbf{looped models for implicit reasoning}, $\mathbf X$ comprises latent slots, recurrently refined under the conditioning of the input prompt, as in LOTUS~\citep{fan2026bridging}.
In \textbf{general looped models}, such as Ouro~\citep{zhu2025scaling} and Huginn~\citep{geiping2025scaling}, $\mathbf X$ can represent context-token inputs whose hidden states are iteratively refined during {recurrent prompt prefilling} to support subsequent decoding.

\paragraph{Adaptive Recurrent Computation.}
We study adaptive computation at the granularity of individual positions, allowing different positions to receive different numbers of recurrent updates.
Specifically, we consider {position-wise freezing}, formally defined as follows.
\begin{definition}[{Position-wise Freezing}]\label{def:freezing}
Freezing position $i$ after pass $\tau_i\in\{0,\ldots,R\}$ halts its subsequent state updates while preserving its layer-wise K/V cache. For every $u>\tau_i$,
\begin{equation}
    \widetilde h_i^{(u)}=\widetilde h_i^{(\tau_i)},\quad
\widetilde\kappa_{i,\ell}^{(u)}
=\widetilde\kappa_{i,\ell}^{(\tau_i)},
\quad \forall \ell,
\end{equation}
where $\widetilde{\cdot}$ denotes quantities under freezing. Active positions continue attending to the preserved K/V throughout subsequent iterations.
\end{definition}

A {freezing schedule} $\boldsymbol{\tau}=(\tau_1,\ldots,\tau_N)$ determines each position's effective recurrent depth $d_i=\tau_i$ and the active width $w_t=|\mathcal A_t|$ at iteration $t$, where $\mathcal A_t={i:\tau_i\ge t}$.
The total number of recurrent position updates is
\begin{equation}
    U(\boldsymbol{\tau})
=\sum_{t=1}^{R}|\mathcal A_t|
=\sum_{i=1}^{N}\tau_i.
\end{equation}
The unfrozen schedule $\boldsymbol{\tau}^{\mathrm{full}}=(R,\ldots,R)$ recovers fixed-loop inference. 
We aim to learn an online freezing policy that reduces recurrent computation while preserving predictive performance.

\paragraph{Freezing Criteria.}
We consider two criteria for adaptive freezing at different levels: {K/V reusability} at the internal representation level, and {functional impact} at the output level.

\begin{definition}[{K/V Reusability}]
\label{def:reuse}
For position $i$ after pass $t<R$, its maximum future K/V deviation along the unfrozen trajectory is
\begin{equation}
    E_i^{\mathrm{KV}}(t)
    = \max_{t<u\le R} D_i^{\mathrm{KV}}(t,u),
\end{equation}
where $D_i^{\mathrm{KV}}$ denotes the layer-averaged
relative K/V distance.
Position $i$ is {reusable} if
$E_i^{\mathrm{KV}}(t)<\epsilon$.
\end{definition}

\begin{definition}[Functional Impact]
\label{def:impact}
Let $\boldsymbol{\tau}^{(i,t)}$ denote the schedule
that freezes only position $i$ after pass $t<R$.
Its {functional impact} is
\begin{equation}
    I_i^{(t)}
    =
    \frac{
        |m(\boldsymbol{\tau}^{(i,t)})
        -m(\boldsymbol{\tau}^{\mathrm{full}})|
    }{
        |m(\boldsymbol{\tau}^{\mathrm{full}})|
    },
\end{equation}
where $a$ and $b$ are the top-two tokens under $\boldsymbol{\tau}^{\mathrm{full}}$ and $m(\boldsymbol{\tau}) = z_{\boldsymbol{\tau}}(a)-z_{\boldsymbol{\tau}}(b)$ denotes the logit margin under schedule $\boldsymbol{\tau}$.
\end{definition}

\section{Empirical Analysis of Recurrent Computation}

\textbf{Setup.}
We analyze the publicly released LOTUS-3B~\cite{fan2026bridging} checkpoint, a looped model for implicit reasoning fine-tuned on GSM8K-Aug~\cite{deng2023implicit}. It refines 150 latent slots over six recurrent loops. 
From the GSM8K validation and test sets, we select 1,648 samples with complete gold reasoning traces to analyze loop requirements. 
We then use correctly solved validation samples for latent-level analyses.

\paragraph{Nonuniform Loop Requirements.}
We investigate computational requirements across both samples and latent positions.
We group samples from the GSM8K validation and test sets by the number of gold reasoning steps $k$ and examine the loop depth sufficient for correct predictions. As shown in Figure~\ref{fig:motiv}a, easier samples, characterized by fewer gold reasoning steps, tend to require fewer loops.
%
At the position level, we examine the earliest loop after which freezing each latent slot individually changes the final prediction margin by less than 1\%.
These freezing depths vary both across and within samples (Figure~\ref{fig:motiv}b).
Together, they highlight the potential for adaptive computation allocation across loop depth and active positions.

\paragraph{Local Convergence and Future Reusability.}
From the internal representation perspective, a key consideration is whether cached K/V remain reusable throughout subsequent computation.
We define a slot as {reusable} after loop $t$ if its cached K/V remain within a tolerance $\epsilon$ along the remaining unfrozen trajectory.
A natural proxy for reusability is local convergence over previous loops.
However, local K/V stability does not necessarily imply future reusability: 
58\% of slot--loop observations exhibiting below-threshold inter-iteration K/V changes subsequently drift beyond $\epsilon$ = 0.45 (Figure~\ref{fig:motiv}c). 
This motivates future-oriented criterion of reusability beyond observed convergence.

\paragraph{Representation Drift and Functional Impact.}
From the output perspective, we consider {functional impact}, which captures the predictive consequences of freezing.
Using exact single-slot counterfactuals, we measure the relative change in the final prediction logit margin when the remaining loops are recomputed with only that slot frozen.
Among slot--loop observations exhibiting substantial future K/V drift, 63\% induce margin changes below 1\% (Figure~\ref{fig:motiv}d).
These findings highlight the distinction between input-level representational consistency and output-level functional impact as two alternative criteria for adaptive freezing.

\paragraph{Inferring Future Utility from Current States.}
\begin{figure}[h]
\centering
  \includegraphics[width=0.95\columnwidth]{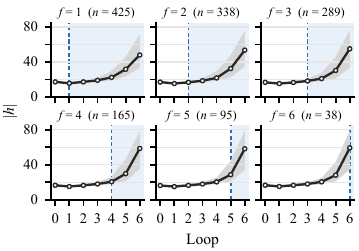}
  \caption{\textbf{Hidden-state norm dynamics across recurrent loops.} 
  }
  \label{fig:motiv_state}
\end{figure}
Both criteria depend on future computation unavailable when freezing decisions are made.
Current hidden states offer a natural source of predictive information, as they feed into subsequent loops.
Yet their observed dynamics do not directly track predictive correctness: across 1,350 GSM8K validation and test instances, hidden-state norms follow similar trajectories regardless of when predictions first become settled, increasing sharply in later loops (Figure~\ref{fig:motiv_state}).
We therefore examine two separately trained lightweight MLPs that estimate K/V reusability and functional impact from current hidden states.
Under five-fold cross-validation on 500 GSM8K validation samples, they achieve within-loop, out-of-fold AUCs of 0.96 and 0.78 on the 372 samples correctly solved after six loops.
These results support learning hindsight-guided freezing decisions from current hidden states.

\section{\textsc{SwiftLoop}}
\label{sec:method}

SwiftLoop instantiates the online freezing policy  through lightweight state-based predictors, each supervised by one of two hindsight-derived signals: {K/V reusability} at the internal representation level and {functional impact} at the output level.
At inference time, the learned policy guides position-wise freezing based on observed hidden states.

\subsection{Online Freezing Policy}
\label{sec:policy}
After each eligible recurrent pass $t<R$, SwiftLoop selectively allocates further refinement by determining which active positions to freeze. 
For each $i\in\mathcal A_t$, we combine its current hidden state and latest recurrent update as
\begin{equation}
\begin{aligned}
\mathbf z_i^{(t)}
&=\left[
h_i^{(t)};\,
h_i^{(t)}-h_i^{(t-1)}
\right]
\end{aligned}
\label{eq:features}
\end{equation}
where $h_i^{(t)}$ is $i$'s hidden state
after pass $t$.
A lightweight predictor $g_\phi$ maps $\mathbf z_i^{(t)}$ to a freezing score $s_i^{(t)}$, with higher scores favoring freezing over continued refinement.
Given a threshold $\lambda$, the policy selects
\begin{equation}
\begin{aligned}
&\mathcal F_t
=\left\{
i\in\mathcal A_t:
s_i^{(t)}\ge\lambda
\right\},\\
&\mathcal A_{t+1}
=\mathcal A_t\setminus\mathcal F_t.
\end{aligned}
\label{eq:freeze-set}
\end{equation}
Each selected position is assigned $\tau_i=t$, skipping all subsequent state and K/V updates while reusing its cached K/V from pass $t$ for attention in later loops.
%
The resulting schedule adaptively
determines the effective depth of
each position and the active width
of each loop. 

\subsection{Hindsight-Guided Policy Learning}
\label{sec:targets}
Both freezing criteria hinge on subsequent recurrent computation, beyond what an online policy can observe. 
SwiftLoop bridges this gap through hindsight-guided learning: supervision is derived offline from full recurrent computation trajectories on calibration sets, enabling online freezing decisions based solely on current hidden states at inference.
We consider two alternative objectives, K/V reusability at the internal representation level and functional impact at the output level, each guiding a separate lightweight freezing predictor.

For each objective, we define a
criterion-specific freezing measure $c_i^{(t)}$
and optimize the lightweight predictor $g_\phi$ on current hidden states:
\begin{equation}
\resizebox{0.95\columnwidth}{!}{$
\mathcal{L}_{\mathrm{reg}}(\phi)
=
\frac{1}{|\mathcal{Q}|}
\sum_{(i,t)\in\mathcal{Q}}
\left(
g_\phi(\mathbf z_i^{(t)})
+
\log(c_i^{(t)}+\delta)
\right)^2,
$}
\label{eq:regression}
\end{equation}
where $\mathcal{Q}$ denotes the
calibration position--pass pairs
and $\delta>0$ ensures numerical
stability.
Higher freezing scores thus
indicate smaller estimated
criterion-specific measures and
favor freezing over continued refinement.

\paragraph{K/V Reusability.}
We derive K/V reusability supervision
from unfrozen $R$-pass reference
trajectories obtained on the
calibration set.
For each position $i$ and eligible pass $t$,
the freezing measure is defined as
the maximum K/V deviation across
subsequent passes:
\begin{equation}
    c_i^{(t)}
    = E_i^{\mathrm{KV}}(t)
    = \max_{t<u\le R}
      D_i^{\mathrm{KV}}(t,u).
\end{equation}

\paragraph{Functional Impact.}
We alternatively characterize freezing
through its effect on the model's output.
Let $a$ and $b$ denote the top two
first-output tokens under the unfrozen
reference run, and define the logit
margin under a freezing schedule
$\boldsymbol{\tau}$ as
$m(\boldsymbol{\tau})
=z_{\boldsymbol{\tau}}(a)
-z_{\boldsymbol{\tau}}(b)$,
with $a$ and $b$ held fixed.

\noindent\underline{Single-position counterfactual}.
Let $\boldsymbol{\tau}^{\mathrm{full}}$
denote the unfrozen schedule, and
$\boldsymbol{\tau}^{(i,t)}$
the schedule that freezes only
position $i$ after pass $t$.
The exact margin change and
corresponding freezing measure are
\begin{equation}
\begin{aligned}
&\Delta_i^{(t)}
=
m(\boldsymbol{\tau}^{(i,t)})
-
m(\boldsymbol{\tau}^{\mathrm{full}}),
\\
&c_{i,\mathrm{exact}}^{(t)}
=
|\Delta_i^{(t)}|.
\end{aligned}
\label{eq:exact_impact_target}
\end{equation}
This directly captures the effect
of an individual freezing decision,
but requires a separate counterfactual
rollout for each position--pass pair.

\noindent\underline{Joint-path IG attribution}.
To obtain scalable supervision,
we attribute the output-level effect
of joint freezing to individual positions
using integrated gradients.
For each position $i$, we introduce
a coefficient $\alpha_i\in[0,1]$
that interpolates its K/V across
subsequent passes:
\begin{equation}
\begin{aligned}
\widetilde{\kappa}_{i,\ell}^{(u)}
(\boldsymbol{\alpha})
=
(1-\alpha_i)
\kappa_{i,\ell}^{(u)}
(\boldsymbol{\alpha})
+
\alpha_i\kappa_{i,\ell}^{(t)},
\quad u>t,
\end{aligned}
\label{eq:kv_interpolation}
\end{equation}
where $\kappa_{i,\ell}^{(u)}
(\boldsymbol{\alpha})$
is recomputed along the
interpolated trajectory.
Let $\mu_t(\boldsymbol{\alpha})$
denote the resulting output margin,
with $\boldsymbol{\alpha}=\mathbf 0$
recovering the unfrozen computation.
We estimate each position's
contribution by integrating its
margin sensitivity along the
joint interpolation path
$\boldsymbol{\alpha}=s\mathbf 1$:
\begin{equation}
\begin{aligned}
\widehat{G}_i^{(t)}
&=
\frac{1}{K}
\sum_{k=1}^{K}
\left.
\frac{\partial\mu_t}
{\partial\alpha_i}
\right|_{\boldsymbol{\alpha}
=\frac{2k-1}{2K}\mathbf 1}
\\
&\approx
\int_0^1
\left.
\frac{\partial\mu_t}
{\partial\alpha_i}
\right|_{\boldsymbol{\alpha}
=s\mathbf 1}
\,ds.
\end{aligned}
\label{eq:ig_impact}
\end{equation}
where $K$ denotes the number
of midpoint quadrature steps.
Each step computes gradients
for all positions in a single backward
pass, enabling efficient functional-impact
estimation compared with position-wise
counterfactual rollouts while capturing interactions
along the joint freezing path.
The resulting freezing measure is
\begin{equation}
c_{i,\mathrm{IG}}^{(t)}
=
|\widehat{G}_i^{(t)}|.
\label{eq:ig_impact_target}
\end{equation}

Both formulations provide alternative
functional-impact targets for
Eq.~\eqref{eq:regression}, measuring
the magnitude of the induced
logit-margin change.


\begin{table*}[t]
\centering
\def\tabdelta#1{{\scriptsize(#1)}}
\resizebox{0.95\textwidth}{!}{%
\begin{tabular}{@{}clr@{\,}lrr@{\,}lrr@{\,}lrr@{\,}lr@{}}
\toprule
\multicolumn{1}{@{}c}{\multirow{2}{*}{\textbf{Model}}} &
\multicolumn{1}{c}{\multirow{2}{*}{\textbf{Method}}} &
\multicolumn{3}{c}{\textbf{GSM8K}} &
\multicolumn{3}{c}{\textbf{SVAMP}} &
\multicolumn{3}{c}{\textbf{MultiArith}} &
\multicolumn{3}{c}{\textbf{GSM-Hard}} \\

\cmidrule(l{2pt}){3-5}
\cmidrule(l{2pt}){6-8}
\cmidrule(l{2pt}){9-11}
\cmidrule(l{2pt}){12-14}

& &
\multicolumn{2}{c}{Acc. (\%)} & Loops &
\multicolumn{2}{c}{Acc. (\%)} & Loops &
\multicolumn{2}{c}{Acc. (\%)} & Loops &
\multicolumn{2}{c}{Acc. (\%)} & Loops \\

\midrule
\multirow{12}{*}{
    \rotatebox[origin=c]{90}{
        \shortstack[c]{LOTUS-1B\\{\footnotesize (CODI-style)}}
    }
} & Fixed (6 loops) & 59.29 & & 6.00 & 61.10 & & 6.00 & 96.67 & & 6.00 & 13.57 & & 6.00 \\
 & Fixed (3 loops) & 56.63 & \tabdelta{$-$2.7} & 3.00 & 60.20 & \tabdelta{$-$0.9} & 3.00 & 97.78 & \tabdelta{+1.1} & 3.00 & 13.04 & \tabdelta{$-$0.5} & 3.00 \\
 & FlashLoop & 59.36 & \tabdelta{+0.1} & 5.86 & 61.20 & \tabdelta{+0.1} & 5.86 & 96.67 & \tabdelta{0.0} & 5.86 & 13.50 & \tabdelta{$-$0.1} & 5.86 \\
 & H2O & 58.61 & \tabdelta{$-$0.7} & 2.75 & 61.00 & \tabdelta{$-$0.1} & 2.75 & 97.22 & \tabdelta{+0.6} & 2.75 & 13.80 & \tabdelta{+0.2} & 2.75 \\
 & LazyLLM & 58.76 & \tabdelta{$-$0.5} & 2.95 & 60.80 & \tabdelta{$-$0.3} & 2.95 & 97.22 & \tabdelta{+0.6} & 2.95 & 13.50 & \tabdelta{$-$0.1} & 2.95 \\
 & CALM & 58.98 & \tabdelta{$-$0.3} & 3.81 & 60.40 & \tabdelta{$-$0.7} & 3.05 & 97.22 & \tabdelta{+0.6} & 3.31 & 13.27 & \tabdelta{$-$0.3} & 3.88 \\
 & GCEE & 58.76 & \tabdelta{$-$0.5} & 3.74 & 60.40 & \tabdelta{$-$0.7} & 3.27 & 97.22 & \tabdelta{+0.6} & 3.32 & 13.27 & \tabdelta{$-$0.3} & 3.71 \\
 & SureLock & 59.21 & \tabdelta{$-$0.1} & 5.89 & 61.20 & \tabdelta{+0.1} & 5.92 & 96.67 & \tabdelta{0.0} & 5.89 & 13.42 & \tabdelta{$-$0.2} & 5.92 \\
 & Huginn KL exit & 59.14 & \tabdelta{$-$0.2} & 5.96 & 61.10 & \tabdelta{0.0} & 5.98 & 96.67 & \tabdelta{0.0} & 5.96 & 13.42 & \tabdelta{$-$0.2} & 5.96 \\
\cmidrule(l{0pt}r{0pt}){2-14}
 & SwiftLoop (reusability) & 58.98 & \tabdelta{$-$0.3} & 3.50 & 61.30 & \tabdelta{$+$0.2} & 1.84 & 97.78 & \tabdelta{$+$1.1} & 1.87 & 13.27 & \tabdelta{$-$0.3} & 3.61 \\
 & SwiftLoop (impact, IG) & 58.83 & \tabdelta{$-$0.5} & 1.80 & 60.10 & \tabdelta{$-$1.0} & 0.79 & 96.67 & \tabdelta{0.0} & 0.98 & 13.42 & \tabdelta{$-$0.2} & 1.52 \\
 & SwiftLoop (impact, exact) & 58.76 & \tabdelta{$-$0.5} & 1.54 & 59.70 & \tabdelta{$-$1.4} & 0.71 & 96.67 & \tabdelta{0.0} & 0.79 & 13.42 & \tabdelta{$-$0.2} & 1.40 \\
\midrule
\multirow{12}{*}{
    \rotatebox[origin=c]{90}{
        \shortstack[c]{LOTUS-3B\\{\footnotesize (CODI-style)}}
    }
} & Fixed (6 loops) & 70.74 & & 6.00 & 75.60 & & 6.00 & 99.44 & & 6.00 & 16.38 & & 6.00 \\
 & Fixed (3 loops) & 39.88 & \tabdelta{$-$30.9} & 3.00 & 71.80 & \tabdelta{$-$3.8} & 3.00 & 97.78 & \tabdelta{$-$1.7} & 3.00 & 9.63 & \tabdelta{$-$6.7} & 3.00 \\
 & FlashLoop & 70.51 & \tabdelta{$-$0.2} & 5.95 & 75.70 & \tabdelta{+0.1} & 5.95 & 99.44 & \tabdelta{0.0} & 5.95 & 16.45 & \tabdelta{+0.1} & 5.95 \\
 & H2O & 71.27 & \tabdelta{+0.5} & 5.40 & 75.20 & \tabdelta{$-$0.4} & 5.40 & 100.00 & \tabdelta{+0.6} & 5.40 & 16.60 & \tabdelta{+0.2} & 5.40 \\
 & LazyLLM & 71.04 & \tabdelta{+0.3} & 5.44 & 75.30 & \tabdelta{$-$0.3} & 5.44 & 100.00 & \tabdelta{+0.6} & 5.44 & 16.76 & \tabdelta{+0.4} & 5.44 \\
 & CALM & 70.96 & \tabdelta{+0.2} & 5.73 & 75.40 & \tabdelta{$-$0.2} & 5.24 & 99.44 & \tabdelta{0.0} & 5.48 & 16.38 & \tabdelta{0.0} & 5.79 \\
 & GCEE  & 70.66 & \tabdelta{$-$0.1} & 5.69 & 75.80 & \tabdelta{+0.2} & 5.49 & 100.00 & \tabdelta{+0.6} & 5.65 & 16.45 & \tabdelta{+0.1} & 5.69 \\
 & SureLock & 70.66 & \tabdelta{$-$0.1} & 5.98 & 75.50 & \tabdelta{$-$0.1} & 5.99 & 99.44 & \tabdelta{0.0} & 5.98 & 16.45 & \tabdelta{+0.1} & 5.98 \\
 & Huginn KL exit & 70.74 & \tabdelta{0.0} & 6.00 & 75.40 & \tabdelta{$-$0.2} & 6.00 & 99.44 & \tabdelta{0.0} & 5.99 & 16.45 & \tabdelta{+0.1} & 6.00 \\
\cmidrule(l{0pt}r{0pt}){2-14}
 & SwiftLoop (reusability) & 71.04 & \tabdelta{+0.3} & 5.65 & 76.00 & \tabdelta{+0.4} & 4.44 & 99.44 & \tabdelta{0.0} & 5.26 & 16.60 & \tabdelta{+0.2} & 5.77 \\
 & SwiftLoop (impact, IG) & 71.11 & \tabdelta{+0.4} & 5.05 & 75.80 & \tabdelta{+0.2} & 3.54 & 100.00 & \tabdelta{+0.6} & 3.97 & 16.38 & \tabdelta{0.0} & 5.20 \\
 & SwiftLoop (impact, exact) & 70.74 & \tabdelta{0.0} & 5.05 & 75.70 & \tabdelta{+0.1} & 3.46 & 98.89 & \tabdelta{$-$0.6} & 3.37 & 16.53 & \tabdelta{+0.2} & 5.46 \\
\midrule
\multirow{12}{*}{
    \rotatebox[origin=c]{90}{
        \shortstack[c]{LOTUS-3B\\{\footnotesize (Coconut-style)}}
    }
} & Fixed (6 loops) & 70.20 & & 6.00 & 73.90 & & 6.00 & 99.44 & & 6.00 & 16.45 & & 6.00 \\
 & Fixed (3 loops) & 54.59 & \tabdelta{$-$15.6} & 3.00 & 71.40 & \tabdelta{$-$2.5} & 3.00 & 98.33 & \tabdelta{$-$1.1} & 3.00 & 13.19 & \tabdelta{$-$3.3} & 3.00 \\
 & FlashLoop & 70.20 & \tabdelta{0.0} & 6.00 & 73.90 & \tabdelta{0.0} & 6.00 & 99.44 & \tabdelta{0.0} & 6.00 & 16.45 & \tabdelta{0.0} & 6.00 \\
 & H2O & 69.14 & \tabdelta{$-$1.1} & 5.74 & 72.60 & \tabdelta{$-$1.3} & 5.74 & 98.89 & \tabdelta{$-$0.6} & 5.74 & 16.07 & \tabdelta{$-$0.4} & 5.74 \\
 & LazyLLM & 69.90 & \tabdelta{$-$0.3} & 5.80 & 73.90 & \tabdelta{0.0} & 5.80 & 99.44 & \tabdelta{0.0} & 5.80 & 16.60 & \tabdelta{+0.2} & 5.80 \\
 & CALM  & 70.20 & \tabdelta{0.0} & 5.85 & 73.90 & \tabdelta{0.0} & 5.44 & 99.44 & \tabdelta{0.0} & 5.69 & 16.30 & \tabdelta{$-$0.2} & 5.87 \\
 & GCEE  & 68.16 & \tabdelta{$-$2.0} & 5.17 & 73.10 & \tabdelta{$-$0.8} & 4.91 & 99.44 & \tabdelta{0.0} & 4.98 & 16.00 & \tabdelta{$-$0.5} & 5.15 \\
 & SureLock & 69.07 & \tabdelta{$-$1.1} & 5.94 & 73.00 & \tabdelta{$-$0.9} & 5.95 & 99.44 & \tabdelta{0.0} & 5.94 & 16.22 & \tabdelta{$-$0.2} & 5.95 \\
 & Huginn KL exit & 69.14 & \tabdelta{$-$1.1} & 5.97 & 73.60 & \tabdelta{$-$0.3} & 5.97 & 99.44 & \tabdelta{0.0} & 5.96 & 16.15 & \tabdelta{$-$0.3} & 5.97 \\
\cmidrule(l{0pt}r{0pt}){2-14}
 & {SwiftLoop} (reusability) & 69.90 & \tabdelta{$-$0.3} & 5.50 & 73.70 & \tabdelta{$-$0.2} & 5.04 & 99.44 & \tabdelta{0.0} & 5.07 & 16.45 & \tabdelta{0.0} & 5.65 \\
 & SwiftLoop (impact, IG) & 70.20 & \tabdelta{0.0} & 5.04 & 73.70 & \tabdelta{$-$0.2} & 3.63 & 99.44 & \tabdelta{0.0} & 3.05 & 16.53 & \tabdelta{+0.1} & 4.52 \\
 & SwiftLoop (impact, exact) & 69.98 & \tabdelta{$-$0.2} & 5.17 & 73.00 & \tabdelta{$-$0.9} & 3.99 & 99.44 & \tabdelta{0.0} & 3.83 & 16.45 & \tabdelta{0.0} & 5.45 \\
\bottomrule
\end{tabular}
}
\caption{\textbf{Main Results}. Loops is the avg. number of loops computed per latent position.}
\label{tab:main-noloss}
\end{table*}

\section{Experiments}
\label{sec:experiments}
\subsection{Experimental Setup}

\paragraph{Models.}
We evaluate SwiftLoop on various looped
models spanning two recurrent
computation settings.
For \textbf{implicit reasoning}, we consider
three LOTUS configurations
\citep{fan2026bridging}, each refining
150 latent positions over six recurrent
loops.
Alongside the released LOTUS-3B
checkpoint, we train CODI-style
LOTUS-1B and LOTUS-3B using
self-distillation
\citep{shen-etal-2025-codi}
on GSM8K-Aug~\citep{deng2023implicit},
initialized from Llama-3.2-1B-Instruct
and Llama-3.2-3B-Instruct, respectively
\citep{grattafiori2024llama}.
For \textbf{recurrent prompt prefilling},
we evaluate Ouro-1.4B
\citep{zhu2025scaling}, which executes
four recurrent steps over 24 layers,
and Huginn-0125~\citep{geiping2025scaling},
a 3.5B model with a four-layer
recurrent block iterated 32 times.

\paragraph{Datasets.}
For implicit reasoning, we evaluate
on the GSM8K test set
\citep{cobbe2021training}
and three out-of-distribution
benchmarks: SVAMP~\citep{patel-etal-2021-nlp},
MultiArith~\citep{roy-roth-2015-solving},
and GSM-Hard~\citep{gao2023pal}.
For recurrent prompt prefilling,
we consider three
benchmarks: WinoGrande
\citep{sakaguchi2021winogrande},
ARC-Challenge~\citep{clark2018think},
and HellaSwag~\citep{zellers-etal-2019-hellaswag}.

\paragraph{Baselines.}
We compare against fixed-depth
inference and a broad range of
adaptive computation methods.
Fixed-depth baselines execute a uniform
number of recurrent steps.
Adaptive baselines include
FlashLoop~\citep{yang2026flashloop},
which allocates computation based on
hidden-state changes;
H2O~\citep{zhang2023ho}
and LazyLLM~\citep{fu2024lazyllm},
which use attention-based importance;
CALM~\citep{schuster2022confident}
and GCEE~\citep{gu2026detecting},
which leverage hidden-state convergence;
SureLock~\citep{oba2026stopping}
and the Huginn KL exit
\citep{geiping2025scaling},
which use output-distribution stability.
More details in Appendix~\ref{app:baselines}.

\paragraph{SwiftLoop Variants.}
We evaluate three variants of SwiftLoop,
sharing the same online freezing
policy and predictor architecture
but differing in their
supervision:
K/V reusability, exact counterfactual
impact, and integrated-gradient
impact.

\paragraph{Evaluation Metrics.}
We report prediction accuracy (\%)
and average recurrent computation.
For implicit reasoning, computation
is measured by the average number
of loops per latent position,
excluding the initialization pass.
For recurrent prompt prefilling,
we report the average number of
recurrent steps per context token.

\paragraph{Implementation Details.}
We fit SwiftLoop predictors on
500 GSM8K validation problems for LOTUS,
and on task-specific training splits
for Ouro and Huginn, with separate
predictors for each model and task.
We consider linear predictors and
one-hidden-layer MLPs with 256 GELU
units, optimized using AdamW with
a learning rate of $10^{-3}$ on
standardized hidden-state features.
Five-fold cross-validation selects
the predictor architecture and
training duration, up to six epochs.
In Eq.~\eqref{eq:regression},
we set $\delta=3\times10^{-5}$
for functional impact in recurrent
prompt prefilling and $10^{-3}$
otherwise.
Integrated-gradient supervision
uses four midpoint quadrature steps.
We select freezing thresholds
to minimize recurrent computation
subject to at most 0.5
percentage point degradation
relative to dense inference.
For LOTUS, thresholds are selected
on the same 500 GSM8K validation
problems used for predictor fitting.
For Ouro and Huginn, we use disjoint
task-specific selection sets.
All thresholds are fixed before
evaluation.
Additional implementation and
calibration details are provided
in Appendix~\ref{app:impl}.

\begin{table}[th]
\centering
\def\tabdelta#1{{\scriptsize(#1)}}
\setlength{\tabcolsep}{3pt}
\resizebox{0.98\columnwidth}{!}{%
\begin{tabular}{@{}lr@{\,}lrr@{\,}lr@{}}
\toprule
\multicolumn{1}{@{}c}{\multirow{2}{*}{\textbf{Method}}} & \multicolumn{3}{c}{\textbf{Ouro-1.4B}} & \multicolumn{3}{c}{\textbf{Huginn-3.5B}} \\
\cmidrule(l{2pt}){2-4} \cmidrule(l{2pt}){5-7}
& \multicolumn{2}{c}{Acc.} & Steps & \multicolumn{2}{c}{Acc.} & Steps \\
\midrule
\multicolumn{7}{@{}l}{\textbf{ARC-C}} \\
\midrule
Fixed (all steps) & 59.98 &  & 4.00 & 44.37 &  & 32.00 \\
FlashLoop & 60.24 & \tabdelta{+0.26} & 2.35 & 42.92 & \tabdelta{$-$1.45} & 19.41 \\
Huginn KL exit & 59.13 & \tabdelta{$-$0.85} & 3.81 & 42.92 & \tabdelta{$-$1.45} & 18.13 \\
SwiftLoop (reusability) & 59.64 & \tabdelta{$-$0.34} & 2.30 & 44.20 & \tabdelta{$-$0.17} & 14.07 \\
SwiftLoop (impact, IG) & 60.49 & \tabdelta{+0.51} & 2.23 & 43.34 & \tabdelta{$-$1.02} & 19.56 \\
\midrule
\multicolumn{7}{@{}l}{\textbf{HellaSwag}} \\
\midrule
Fixed (all steps) & 74.37 &  & 4.00 & 67.97 &  & 32.00 \\
FlashLoop & 74.37 & \tabdelta{0.00} & 2.35 & 67.58 & \tabdelta{$-$0.40} & 19.42 \\
Huginn KL exit & 74.41 & \tabdelta{+0.04} & 3.98 & 67.90 & \tabdelta{$-$0.07} & 19.79 \\
SwiftLoop (reusability) & 74.65 & \tabdelta{+0.28} & 2.47 & 67.84 & \tabdelta{$-$0.14} & 12.36 \\
SwiftLoop (impact, IG) & 74.44 & \tabdelta{+0.07} & 2.01 & 67.62 & \tabdelta{$-$0.36} & 19.27 \\
\midrule
\multicolumn{7}{@{}l}{\textbf{WinoGrande}} \\
\midrule
Fixed (all steps) & 70.88 &  & 4.00 & 63.22 &  & 32.00 \\
FlashLoop & 70.72 & \tabdelta{$-$0.16} & 2.36 & 62.98 & \tabdelta{$-$0.24} & 19.50 \\
Huginn KL exit & 70.88 & \tabdelta{0.00} & 3.99 & 62.12 & \tabdelta{$-$1.10} & 20.23 \\
SwiftLoop (reusability) & 69.69 & \tabdelta{$-$1.18} & 3.34 & 63.14 & \tabdelta{$-$0.08} & 13.99 \\
SwiftLoop (impact, IG) & 70.40 & \tabdelta{$-$0.47} & 2.05 & 62.90 & \tabdelta{$-$0.32} & 18.95 \\
\bottomrule
\end{tabular}}
\caption{{Generalization to Recurrent Prompt Prefilling}.}
\label{tab:prefill-noloss}
\end{table}

\begin{figure*}[t]
\centering
  \includegraphics[width=0.99\textwidth]{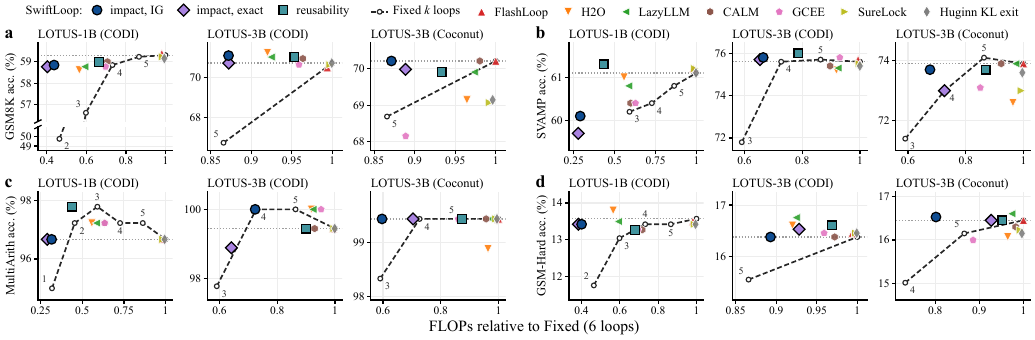}
  \caption{Accuracy versus FLOPs relative to Fixed (6 loops).}
  \label{fig:flop}
\end{figure*}

\subsection{Main Results on Implicit Reasoning}
\label{sec:exp-reasoning}

\paragraph{Selective refinement improves the accuracy–efficiency trade-off.}
Table~\ref{tab:main-noloss} presents
the accuracy--computation trade-offs
across reasoning benchmarks,
with the corresponding FLOPs comparisons
shown in Figure~\ref{fig:flop}.
Uniformly halving recurrent depth
reduces GSM8K accuracy by 2.7, 30.9,
and 15.6 points across the three
LOTUS models, highlighting the
limitations of fixed computation
budgets.
In contrast, SwiftLoop achieves
substantial savings through selective
position-wise freezing.
On CODI-style LOTUS-1B,
SwiftLoop (impact, IG) achieves
58.83\% accuracy with only 1.80
loops per latent, reducing recurrent
computation by 70\%.
On the two 3B models, it preserves
or slightly improves GSM8K accuracy while reducing recurrent computation.
Across all 12 model--benchmark pairs,
SwiftLoop requires
fewer loops than every adaptive
baseline, with accuracy remaining
within 0.5 points of dense inference
in 11 settings.
At comparable FLOPs, it also
outperforms fixed-loop inference
by 9.1, 4.4, and 1.5 accuracy
points on GSM8K across the
three models, respectively.

\paragraph{Advantages across baselines and benchmarks.}
Existing adaptive methods generally
offer less favorable trade-offs.
FlashLoop, SureLock, and the Huginn
KL exit retain nearly all recurrent
computation, while CALM and GCEE
achieve limited savings on the
3B models.
Attention-based methods H2O and
LazyLLM are more competitive on
LOTUS-1B, requiring 2.75--2.95 loops,
but still execute 5.40--5.80 loops
on the 3B models.
SwiftLoop's advantages extend to
out-of-distribution benchmarks.
On LOTUS-1B, K/V reusability
achieves 61.30\% accuracy on SVAMP
and 97.78\% on MultiArith with
fewer than two loops.
On CODI-style LOTUS-3B,
SwiftLoop (impact, IG) achieves
75.80\% accuracy on SVAMP with
3.54 loops and 100\% on MultiArith
with 3.97 loops.
These results demonstrate effective
computation reduction across model
scales, training recipes, and
reasoning benchmarks.

\begin{figure*}[t]
\centering
  \includegraphics[width=0.9\textwidth]{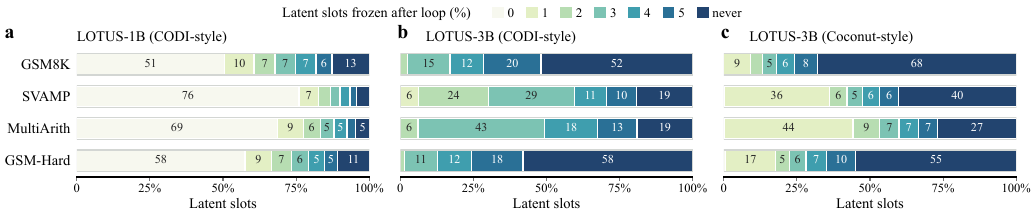}
  \caption{Share of latent slots frozen after each loop under SwiftLoop (impact, IG).}
  \label{fig:break}
\end{figure*}

\begin{figure}[h]
\centering
  \includegraphics[width=0.95\columnwidth]{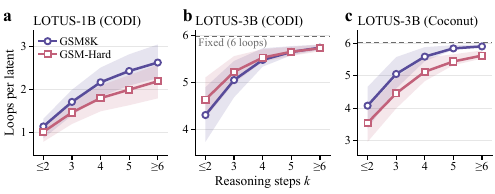}
  \caption{Adaptive Loop Allocation across Problems.}
  \label{fig:diffloop}
\end{figure}

\subsection{Generalization to Recurrent Prompt Prefilling}
\label{sec:exp-prefill}
We further evaluate SwiftLoop on recurrent prompt prefilling, applying position-wise freezing to context tokens in Ouro and Huginn.
Table~\ref{tab:prefill-noloss}
shows that SwiftLoop substantially
reduces recurrent computation
across Ouro and Huginn while
largely preserving prediction accuracy.

On Ouro-1.4B, integrated-gradient supervision
achieves the most favorable
accuracy--computation trade-off,
requiring only 2.01--2.23 steps
per context token.
In comparison, FlashLoop requires
2.35--2.36 steps at similar accuracy,
while the Huginn KL exit retains
3.81--3.99 of the four steps.
K/V reusability is less effective
on Ouro, particularly on WinoGrande,
where it incurs a 1.18-point
accuracy drop while using
3.34 steps.
On Huginn-3.5B, K/V reusability provides stronger
accuracy--computation trade-offs, reducing the average
recurrent steps from 32 to
12.36--14.07, a 56--61\% reduction.
Across all three benchmarks,
accuracy remains within 0.17
percentage points of full recurrence.
FlashLoop and the Huginn KL exit
require 18.13--20.23 steps,
with accuracy losses reaching
1.45 points.
Integrated-gradient supervision
requires 18.95--19.56 steps,
yielding savings comparable to
FlashLoop.
These results demonstrate SwiftLoop's applicability across recurrent architectures, with representation-level K/V reusability and output-level functional impact offering complementary perspectives on selective refinement.

\subsection{Inference-Time Adaptive Computation}
\label{sec:exp-analysis}

SwiftLoop enables fine-grained adaptive
computation by dynamically adjusting
the effective depth of individual
positions and the active width of
each recurrent loop.
Figure~\ref{fig:break} reveals
heterogeneous computation allocation
across models and benchmarks.
LOTUS-1B freezes 51--76\% of
latent positions immediately after
initialization, while the two
3B models exhibit distinct
freezing patterns across
recurrent loops.
This adaptivity further emerges
at the problem level.
Figure~\ref{fig:diffloop} shows that
average loops per latent increase
consistently with reference reasoning
length across all three models on
GSM8K and GSM-Hard.
Notably, these patterns emerge
without explicit difficulty supervision,
while variations within each difficulty
group indicate instance-specific
computation allocation.
Together, these findings demonstrate
how local freezing decisions enable
adaptive computation across both
positions and samples.

\begin{table}[t]
\centering
\setlength{\tabcolsep}{2.5pt}
\resizebox{0.98\columnwidth}{!}{%
\begin{tabular}{@{}lcccccc@{}}
\toprule
\multicolumn{1}{@{}l}{\multirow{2}{*}{\textbf{Implicit Reasoning}}} & \multicolumn{2}{c}{\textbf{GSM8K}} & \multicolumn{2}{c}{\textbf{SVAMP}} & \multicolumn{2}{c}{\textbf{MultiArith}} \\
\cmidrule(l{2pt}){2-3} \cmidrule(l{2pt}){4-5} \cmidrule(l{2pt}){6-7}
& Reus. & IG & Reus. & IG & Reus. & IG \\
\midrule
LOTUS-1B (CODI) & 1.33$\times$ & 1.76$\times$ & 1.93$\times$ & 2.26$\times$ & 1.93$\times$ & 2.48$\times$ \\
LOTUS-3B (CODI) & 1.01$\times$ & 1.09$\times$ & 1.20$\times$ & 1.38$\times$ & 1.07$\times$ & 1.29$\times$ \\
LOTUS-3B (Coconut) & 1.01$\times$ & 1.07$\times$ & 1.08$\times$ & 1.33$\times$ & 1.08$\times$ & 1.50$\times$ \\
\midrule
\multicolumn{1}{@{}l}{\multirow{2}{*}{\textbf{Prompt Prefilling}}} & \multicolumn{2}{c}{\textbf{WinoGrande}} & \multicolumn{2}{c}{\textbf{ARC-C}} & \multicolumn{2}{c}{\textbf{HellaSwag}} \\
\cmidrule(l{2pt}){2-3} \cmidrule(l{2pt}){4-5} \cmidrule(l{2pt}){6-7}
& Reus. & IG & Reus. & IG & Reus. & IG \\
\midrule
Ouro-1.4B & 1.12$\times$ & 1.43$\times$ & 1.56$\times$ & 1.62$\times$ & 1.52$\times$ & 1.86$\times$ \\
Huginn-3.5B & 1.82$\times$ & 1.48$\times$ & 1.46$\times$ & 1.26$\times$ & 1.47$\times$ & 1.26$\times$ \\
\bottomrule
\end{tabular}}
\caption{\textbf{Measured Speedup}. Wall-clock speedup of SwiftLoop relative to dense inference.}
\label{tab:latency-all}
\end{table}

\subsection{Wall-Clock Speedup}

We evaluate the practical inference
efficiency of SwiftLoop through
end-to-end wall-clock measurements
(Table~\ref{tab:latency-all}).
For implicit reasoning, IG-supervised
SwiftLoop achieves $1.76\times$,
$2.26\times$, and $2.48\times$
speedups on LOTUS-1B across GSM8K,
SVAMP, and MultiArith, respectively.
On the two LOTUS-3B models, it delivers
$1.07$--$1.50\times$ speedups
across these benchmarks.
SwiftLoop also accelerates recurrent
prompt prefilling, achieving
$1.43$--$1.86\times$ speedups
with functional-impact supervision
on Ouro and $1.46$--$1.82\times$
with K/V reusability on Huginn.
Notably, Huginn achieves
$1.46\times$ and $1.47\times$
speedups on ARC-Challenge and
HellaSwag, respectively, under
single-context prefilling.
These results demonstrate
SwiftLoop's practical inference
efficiency across implicit reasoning
and recurrent prompt prefilling.


\section{Related Work}
\noindent\textbf{Implicit Reasoning}.
Implicit chain-of-thought reasoning replaces explicit reasoning traces with continuous latent-state computation~\citep{deng2023implicit,deng2024explicit}.
Prior work advances implicit reasoning through diverse strategies, including
pause tokens~\citep{goyal2024think}, curriculum-based latent reasoning \citep{hao2025training},
self-distillation \citep{shen-etal-2025-codi},
reasoning compression
\citep{cheng2024compressed,xu-etal-2025-softcot},
and semantic or step-level supervision
\citep{he2025semcot,wei2026simcot}.
PCCoT~\citep{wu-etal-2025-parallel}
explores parallel latent reasoning,
while LOTUS~\citep{fan2026bridging}
builds on a looped Transformer
architecture to iteratively refine
latent slots for
implicit reasoning.

\noindent\textbf{Looped Transformers and Adaptive Computation}.
Looped Transformers increase effective
computational depth through repeated
application of shared modules
\citep{dehghani2018universal}.
Their capabilities for algorithmic and multi-step reasoning have been extensively studied~\citep{giannou2023looped,yang2024looped,saunshi2025reasoning}.
Recent recurrent-depth language models,
including Huginn~\citep{geiping2025scaling}
and Ouro~\citep{zhu2025scaling},
further demonstrate the potential
of iterative latent computation.
The computational demands of recurrence
have inspired efficient inference
through K/V cache optimization
\citep{wu2025parallel,vendrell2026memory},
efficient prefilling~\citep{wan2026attention,du2026not},
and accelerated decoding
\citep{yu-etal-2026-loopcd,liu2026decoding,cho2026loopspec}.
Adaptive computation further improves
recurrent efficiency by allocating
computation across inputs and positions.
Existing approaches learn adaptive depth
through dynamic halting and token-wise
routing~\citep{graves2016adaptive,
banino2021pondernet,
bae2025mixtureofrecursions,
fu2026thinkathard,wang2026recurtrace},
while inference-time methods guide
selective updates through attention
importance, hidden-state dynamics,
and output-distribution stability
\citep{zhang2023ho,fu2024lazyllm,
schuster2022confident,gu2026detecting,
yang2026flashloop,oba2026stopping,
geiping2025scaling}.
SwiftLoop learns position-wise
freezing decisions through
hindsight-guided supervision,
considering future K/V reusability
at the internal representation level
and functional impact at the
output level.


\section{Conclusion}
We present \textsc{SwiftLoop},
a hindsight-guided framework for
selective refinement in looped Transformers.
Through position-wise freezing and
K/V reuse, SwiftLoop adaptively
allocates computation across recurrent
depth and active width.
Lightweight predictors learn from
hindsight supervision based on
representation-level K/V reusability
or output-level functional impact,
and guide adaptive freezing using
only current hidden states at inference time.
Extensive experiments demonstrate favorable accuracy–computation trade-offs and practical inference acceleration in implicit reasoning, with further generalization to recurrent prompt prefilling.






\bibliography{custom}

\appendix

\section{K/V Distance Definition}
\label{app:measure-impl}

For position $i$, let
$\mathbf k_i^{(\ell)}(t)$ and
$\mathbf v_i^{(\ell)}(t)$ denote
the key and value vectors at layer
$\ell$ and pass $t$, with all
K/V heads concatenated.
We define
$\mathbf z_i^{(\ell)}(t)
=[\mathbf k_i^{(\ell)}(t);
\mathbf v_i^{(\ell)}(t)]$
and compute the layer-averaged
relative distance as
\begin{equation}
D_i^{\mathrm{KV}}(t,u)
=
\frac{1}{L}
\sum_{\ell=1}^{L}
\frac{
\|\mathbf z_i^{(\ell)}(t)
-\mathbf z_i^{(\ell)}(u)\|_2
}{
\max(
\|\mathbf z_i^{(\ell)}(u)\|_2,
10^{-6}
)
}.
\label{eq:kv-distance-app}
\end{equation}
Here, $L$ denotes the recurrent
layers.
The normalization uses the K/V
norm at the later pass $u$.
Distances are measured at the
K/V projection outputs; rotary
position embeddings preserve
these distances across passes.
The reusability target
$E_i^{\mathrm{KV}}(t)
=\max_{u>t}D_i^{\mathrm{KV}}(t,u)$
is computed from unfrozen
calibration trajectories.

\section{Experimental Details}
\label{app:impl}

This section provides additional details
on the model configurations, evaluation
benchmarks, predictor fitting, and
experimental protocols used in
Section~\ref{sec:experiments}.

\subsection{Models and Benchmarks}

\paragraph{LOTUS Models.}
We evaluate three LOTUS configurations
\citep{fan2026bridging}, covering
different model scales and
implicit-reasoning training recipes.
All three models use a recurrent
workspace of 150 latent positions
and perform six recurrent refinement
loops following an initialization pass.
The first configuration is the released
LOTUS-3B checkpoint, which follows the
Coconut-style post training adaptation.
To assess SwiftLoop across different
training recipes, we additionally
train two CODI-style LOTUS models
using the self-distillation approach
of CODI~\citep{shen-etal-2025-codi}
on GSM8K-Aug~\citep{deng2023implicit}.
The 1B and 3B configurations are
initialized from
Llama-3.2-1B-Instruct and
Llama-3.2-3B-Instruct, respectively
\citep{grattafiori2024llama}.
All three configurations share the
same recurrent workspace structure
and freezing interface.
Under full six-loop inference,
CODI-style LOTUS-1B,
CODI-style LOTUS-3B, and
Coconut-style LOTUS-3B achieve
59.29\%, 70.74\%, and 70.20\%
accuracy on GSM8K test, respectively.

\paragraph{General Looped Models.}
We further evaluate SwiftLoop on
Ouro-1.4B~\citep{zhu2025scaling}
and Huginn-0125~\citep{geiping2025scaling},
which perform recurrent computation
over prompt-token representations.
Ouro-1.4B has 24 layers and executes
four recurrent steps.
Huginn-0125 contains approximately
3.5B parameters and applies a
four-layer recurrent block for
32 steps, preceded by a two-layer
prelude and followed by a two-layer
coda.
Unlike LOTUS, where recurrence
refines latent reasoning positions,
these models recurrently process
context-token states.
SwiftLoop applies the same
position-wise freezing rule to
eligible context positions,
preserving their cached K/V for
attention in subsequent steps.
We enable adaptive freezing after
two recurrent steps for Ouro
and four for Huginn, following
the warm-up convention of
FlashLoop~\citep{yang2026flashloop}.
Huginn's stochastic initial state
is generated using a fixed seed,
ensuring identical initialization
across evaluated methods.

\paragraph{Implicit-Reasoning Benchmarks.}
We evaluate mathematical reasoning
on four benchmarks:

\begin{itemize}
\item \textbf{GSM8K}
\citep{cobbe2021training}:
1,319 test problems involving
multi-step grade-school arithmetic
and mathematical reasoning.

\item \textbf{SVAMP}
\citep{patel-etal-2021-nlp}:
1,000 mathematical word problems
designed to evaluate robustness
to variations in problem structure
and linguistic formulation.

\item \textbf{MultiArith}
\citep{roy-roth-2015-solving}:
180 multi-step arithmetic
word problems.

\item \textbf{GSM-Hard}
\citep{gao2023pal}:
1,319 challenging numerical
variations of GSM8K problems.
\end{itemize}
GSM8K serves as the primary
in-distribution benchmark, while
SVAMP, MultiArith, and GSM-Hard
evaluate out-of-distribution
generalization.
All models produce answers through
greedy decoding after recurrent
computation.
We extract the numerical answer
and compare it with the reference
answer for correctness.

\paragraph{Prompt-Prefilling Benchmarks.}
We evaluate recurrent prompt
prefilling on three
benchmarks:

\begin{itemize}
\item \textbf{WinoGrande}
\citep{sakaguchi2021winogrande}:
a commonsense reasoning benchmark
requiring resolution of ambiguous
references.
We use five-shot evaluation on
1,267 validation instances.

\item \textbf{ARC-Challenge}
\citep{clark2018think}:
multiple-choice science questions
requiring nontrivial reasoning
and background knowledge.
We use 25-shot evaluation on
1,172 test instances.

\item \textbf{HellaSwag}
\citep{zellers-etal-2019-hellaswag}:
a commonsense inference benchmark
requiring selection of plausible
event continuations.
We use ten-shot evaluation on
10,042 validation instances.
\end{itemize}
Following the evaluation protocol
of lm-evaluation-harness
\citep{biderman2024lessons},
we score each candidate continuation
by its conditional log-likelihood
and select the highest-scoring
candidate.
For ARC-Challenge and HellaSwag,
candidate scores are normalized
by continuation length.
The same prompting and scoring
protocol is used across methods
for each backbone and benchmark.

\subsection{Implementation Details}

\paragraph{Freezing and Feature Extraction.}
SwiftLoop applies irreversible
position-wise freezing while
retaining cached K/V for attention
by active positions.
For LOTUS, the initialization pass
is indexed by $t=0$ and the six
recurrent loops by $t=1,\ldots,6$.
Freezing decisions may be made
after passes $t=0,\ldots,5$.
The predictor input concatenates
the current hidden state with
its most recent update, as
defined in Eq.~\eqref{eq:features}.
For the initial pass, where
no preceding recurrent state
is available, we set the
update component to zero.
For recurrent prompt prefilling,
freezing operates on context-token
positions.
Frozen positions retain their hidden
states and cached K/V.
The active-position set decreases
monotonically over recurrent steps.

\paragraph{Hindsight Supervision.}
For each backbone, we collect
reference recurrent trajectories
to construct the supervision
targets described in
Section~\ref{sec:targets}.
For K/V reusability, we use the
maximum K/V deviation across
subsequent recurrent passes:
\begin{equation}
c_i^{(t)}
=
E_i^{\mathrm{KV}}(t)
=
\max_{t<u\le R}
D_i^{\mathrm{KV}}(t,u).
\end{equation}
These targets are obtained
from unfrozen reference
trajectories.
For exact functional impact,
we freeze one position after
pass $t$, retain full recurrence
at all other positions, and
recompute the resulting output
margin.
The freezing measure is the
absolute change relative to
the unfrozen reference margin:
\begin{equation}
c_{i,\mathrm{exact}}^{(t)}
=
\left|
m(\boldsymbol{\tau}^{(i,t)})
-
m(\boldsymbol{\tau}^{\mathrm{full}})
\right|.
\end{equation}
For integrated-gradient supervision,
we interpolate recurrent K/V
along the joint freezing path
introduced in
Section~\ref{sec:targets}.
We approximate the path integral
using midpoint quadrature,
with four integration points
for LOTUS and two for Huginn.
Each point provides gradient
attributions for all eligible
positions through a shared
forward--backward computation.
For output-level supervision,
the reference margin uses the
top two first-output tokens
of the unfrozen computation,
with their identities held
fixed during evaluation.

\paragraph{Predictor Fitting.}
We fit separate freezing predictors
for each backbone and supervision
objective.
For recurrent prompt prefilling,
predictors are additionally
fitted separately for each task.
Each predictor receives the
concatenated hidden-state features
defined in Eq.~\eqref{eq:features}.
We consider a linear predictor
and a one-hidden-layer MLP
with 256 GELU units.
Input features are standardized,
and both architectures are
optimized using AdamW with
learning rate $10^{-3}$.
We select the predictor architecture
and training duration using
five-fold cross-validation,
partitioned by problem or document
to keep observations from the
same input within a single fold.
Training is limited to six epochs.
The selected configuration is
subsequently fitted for online use.
Both supervision objectives use
the log-scale regression loss
in Eq.~\eqref{eq:regression}.
We set $\delta=3\times10^{-5}$
for functional impact in recurrent
prompt prefilling, and $10^{-3}$
otherwise.
Higher predicted freezing scores
correspond to smaller estimated
freezing measures.

For LOTUS, predictors are fitted
using 500 GSM8K validation problems.
For Ouro and Huginn, supervision
is constructed from the training
split of the corresponding task,
with separate predictors fitted
for each model--task combination.

\paragraph{Threshold Selection.}
We select the online freezing
threshold from a candidate grid
by evaluating the resulting
freezing schedules.
For each configuration, we choose
the threshold minimizing average
recurrent computation subject to
a prescribed accuracy tolerance
relative to dense inference.
We consider \emph{0.5-point loss}, where a
validation accuracy degradation of at most 0.5
percentage point is permitted.
When no candidate satisfies the
constraint, we retain the
highest-accuracy configuration.

For LOTUS, threshold selection
uses the same 500 GSM8K validation
problems used for predictor fitting.
The selected threshold is applied
unchanged to GSM8K test, SVAMP,
MultiArith, and GSM-Hard.
For recurrent prompt prefilling,
we use task-specific selection
instances disjoint from the
predictor-fitting inputs.
The selection sets contain
600 WinoGrande, 719 ARC-Challenge,
and 600 HellaSwag training-split
instances, respectively.
No selection uses the final
evaluation instances.

All thresholds are fixed before
evaluation.
The chosen operating points
are determined from full-policy
executions, thereby reflecting
the joint effect of freezing
multiple positions.

\paragraph{Computational Resources.}
All methods use the same
checkpoint and evaluation protocol
within each model configuration.
Experiments were conducted using NVIDIA A100 GPUs.

\subsection{Training Recipe of CODI-Style LOTUS Models}
\label{app:codi-recipe}

\begin{table}[h]
\centering
\resizebox{0.99\columnwidth}{!}{%
\begin{tabular}{@{}lcc@{}}
\toprule
\textbf{Configuration} &
\textbf{LOTUS-1B (CODI)} &
\textbf{LOTUS-3B (CODI)} \\
\midrule
Initialization & Llama-3.2-1B-Instruct & Llama-3.2-3B-Instruct \\
LoRA rank / $\alpha$ / dropout & \multicolumn{2}{c}{128 / 32 / 0.1} \\
Trainable parameters & 90.2M & 194.5M \\
Learning rate & $8\times10^{-4}$ & $3\times10^{-4}$ \\
Epochs (steps) & 10 (29,990) & 8 (23,992) \\
Batch / GPU & 16 & 8 \\
Gradient accumulation & 4 & 8 \\
Global batch size & \multicolumn{2}{c}{128} \\
Optimizer / schedule & \multicolumn{2}{c}{AdamW / cosine (3\% warm-up)} \\
Weight decay / grad. clipping & \multicolumn{2}{c}{0.1 / 2.0} \\
Training time & $\sim$45 h & $\sim$96 h \\
\bottomrule
\end{tabular}}
\caption{\textbf{Training configurations of CODI-style LOTUS models.}
Training uses two A100 80GB GPUs with bfloat16 precision.}
\label{tab:codi-recipe}
\end{table}

\noindent\textbf{Architecture.}
We train CODI-style LOTUS-1B and LOTUS-3B
from Llama-3.2-1B-Instruct and
Llama-3.2-3B-Instruct, respectively,
following the LOTUS architecture.
Both models refine 150 latent slots,
organized into six blocks of 25,
over six recurrent loops.
Each loop incorporates the preceding
hidden states into the slot embeddings,
while reusing the question's K/V cache.
Training backpropagates through all
seven passes (initialization and six loops),
without a recurrent-depth curriculum.
The added token embeddings remain fixed,
and only LoRA adapters are optimized.

\noindent\textbf{Training Objective.}
Following CODI, a shared-parameter model
serves as both the latent student and
the explicit-CoT teacher.
We optimize
\begin{equation}
\mathcal{L}
=
\mathcal{L}_{\mathrm{ans}}
+
\lambda_{\mathrm{dist}}\mathcal{L}_{\mathrm{dist}}
+
\mathcal{L}_{\mathrm{teacher}}
+
\lambda_{\mathrm{read}}\mathcal{L}_{\mathrm{read}},
\end{equation}
where $\lambda_{\mathrm{dist}}=20$
and $\lambda_{\mathrm{read}}=0.05$.
The student predicts the final answer
after recurrent refinement, while the
teacher receives explicit CoT supervision.
We align their hidden states at the
answer-prediction position using
teacher-normalized SmoothL1 loss
across embedding, Transformer, and
final-normalization layers, with
teacher representations detached.
To integrate CODI self-distillation
with parallel latent reasoning,
we additionally apply LOTUS's
token-level supervision:
the final-pass slots in block $b$
predict the tokens of reasoning step $b$.

\noindent\textbf{Data and Optimization.}
We train on 383,820 GSM8K-Aug problems
after CODI-style filtering.
Table~\ref{tab:codi-recipe} summarizes
the LoRA configuration, optimization
hyperparameters, and training schedules.
Both models use AdamW with cosine decay,
a global batch size of 128, and
bfloat16 precision.
Training uses two NVIDIA A100 80GB GPUs,
taking approximately 45 hours for
LOTUS-1B and 96 hours for LOTUS-3B.
The final checkpoints achieve
59.3\% and 70.7\% GSM8K accuracy,
respectively, under full recurrence.

\section{Baselines}
\label{app:baselines}

We describe the baseline methods
used in our experiments, including
their original decision criteria,
adaptations to recurrent computation,
and configuration procedures.

\noindent\textbf{Common Protocol.}
For adaptive baselines originally
designed for other inference settings,
we adapt their selection criteria
to position-wise recurrent computation.
In LOTUS, freezing operates on
latent reasoning positions.
In Ouro and Huginn, it operates
on context-token positions.
Frozen positions retain their
cached K/V for attention by
subsequently updated positions.
Unless otherwise specified,
baseline configurations are fixed
before evaluation.
H2O and LazyLLM use the same
accuracy-constrained selection
rule as SwiftLoop, while methods
with established calibration
procedures or published defaults
retain their respective settings.
Detailed baseline configurations and
calibration procedures are described
below.

\subsection{Fixed-Depth Baselines}

We consider uniform recurrent
computation with a fixed number
of loops $K$ across positions.
For LOTUS, $K=6$ correspondes
to full recurrence.
For Ouro and Huginn, full
recurrence comprises 4 and
32 steps, respectively, and
reduced-step inference uses
smaller recurrent budgets.

\subsection{Adaptive Baselines}

\paragraph{FlashLoop.}
FlashLoop~\citep{yang2026flashloop}
reduces recurrent computation
by prioritizing positions with
the largest hidden-state changes.
Its retention schedules are
calibrated against the hidden-state
reconstruction error of
dense recurrence.
For LOTUS, we measure the
normalized recurrent update
of each active latent:
\begin{equation}
q_i^{(t)}
=
\frac{
\|h_i^{(t)}-h_i^{(t-1)}\|_2
}{
\|h_i^{(t)}\|_2+10^{-6}
}.
\end{equation}
At each eligible pass, we retain
the prescribed number of positions
with the largest updates and
freeze the remaining positions.
The retained sets are nested
across subsequent loops.
The relative reconstruction error is
\begin{equation}
e_t
=
\frac{
\|H_{\mathrm{adaptive}}^{(t)}
-H_{\mathrm{dense}}^{(t)}\|_F
}{
\|H_{\mathrm{dense}}^{(t)}\|_F
}.
\end{equation}
We search retention shares on
a grid with 5\% increments,
selecting the smallest shares
that maintain an average
reconstruction error below 5\%,
subject to non-increasing
retention across loops.
For recurrent prompt prefilling,
we use FlashLoop's reported
backbone-specific retention
schedules.

\paragraph{H2O.}
H2O~\citep{zhang2023ho}
was originally proposed for
efficient generative inference
through KV-cache management.
It prioritizes heavy-hitter
tokens receiving high accumulated
attention while retaining
recent tokens.
We adapt its attention-based
ranking to recurrent positions,
replacing cache eviction with
position-wise freezing.
For LOTUS, a latent's importance
score accumulates the attention
received from active latent
queries across recurrent passes,
averaged over heads and layers.
To account for causal attention
asymmetry, we normalize by
the number of active queries
that can attend to each position.
After a selected pass $s$,
a fraction $r$ of positions
remains active.
Half are selected by accumulated
attention scores and half
by recency, while the
remaining positions are frozen.
For LOTUS, we search 23 candidate
configurations with
$0.2\le r\le0.95$ and $s\le3$.
Attention scores are obtained
through an auxiliary
eager-attention evaluation,
whose overhead is excluded from
the reported recurrent-loop count.

\paragraph{LazyLLM.}
LazyLLM~\citep{fu2024lazyllm}
performs dynamic token pruning
during long-context inference,
using attention-based importance
to identify positions requiring
further computation.
We adapt its selection rule
to recurrent positions while
preserving cached K/V after
freezing.
For LOTUS, latent importance
is measured by the attention
received from the final token
of the answer prompt, averaged
over heads and layers.
After each eligible pass
$t\ge s$, we freeze the fraction
$p$ of active latents with
the lowest attention scores.
We search 25 candidate
$(p,s)$ configurations,
with $0.02\le p\le0.6$
and $s\le3$, selecting
the operating point on
GSM8K validation.
The auxiliary answer-prompt
evaluation incurs
0.06--0.24 loop-equivalents
per pass on GSM8K, depending
on the backbone.
%

\paragraph{CALM.}
CALM~\citep{schuster2022confident}
enables adaptive early exit
through hidden-state convergence.
We adopt its saturation criterion,
which measures the cosine similarity
between consecutive representations.
For LOTUS, latent $i$ is frozen
after pass $t$ when
\begin{equation}
\cos\left(
h_i^{(t)},h_i^{(t-1)}
\right)\ge\lambda.
\end{equation}
The frozen position retains
its cached K/V for
subsequent attention.
We calibrate $\lambda$ using
CALM's Learn-then-Test procedure
\citep{angelopoulos2025learn}
on GSM8K validation, with
risk tolerance $\delta=0.02$
and significance level
$\epsilon=0.05$.
The risk indicator is one
when dense inference answers
correctly but the adaptive
configuration does not,
and zero otherwise.
Candidate thresholds are obtained
from quantiles of reference
cosine similarities.
We apply fixed-sequence testing
with Hoeffding--Bentkus
$p$-values to select the
most aggressive threshold
accepted by the calibration
procedure.

\paragraph{GCEE.}
GCEE~\citep{gu2026detecting}
detects convergence in
hidden-state trajectories
using both update magnitude
and directional stability.
We apply its criteria to
individual LOTUS latent
positions across recurrent
passes.
Let
$u_i^{(t)}
=h_i^{(t)}-h_i^{(t-1)}$.
A position is frozen when
\begin{equation}
\begin{aligned}
&\frac{\|u_i^{(t)}\|_2}
{\|h_i^{(t)}\|_2}
\le\tau_s,
\\
&\cos\left(
u_i^{(t)},u_i^{(t-1)}
\right)
\ge\tau_c
\end{aligned}
\end{equation}
hold for two consecutive passes.
The earliest possible freezing
decision is therefore after
pass $t=3$.
Following GCEE's calibration
approach, we search
$\tau_c\in\{0,0.5,0.9\}$
and $\tau_s$ values derived
from quantiles of the
reference update magnitudes.
We select the configuration
with the fewest recurrent
updates that retains at least
99\% of dense validation
accuracy.

\paragraph{SureLock.}
SureLock~\citep{oba2026stopping}
identifies converged token
representations in masked
diffusion language models
through changes in their
output distributions.
For LOTUS, we apply the LM head
to each latent hidden state:
\begin{equation}
p_i^{(t)}
=
\operatorname{softmax}
\left(
\operatorname{LMHead}(h_i^{(t)})
\right).
\end{equation}
A latent is frozen when
\begin{equation}
D_{\mathrm{KL}}
\left(
p_i^{(t)}
\Vert p_i^{(t-1)}
\right)
\le\epsilon
\end{equation}
and its prediction uncertainty
lies within the lowest 20\%
among active positions.
We use the published
configuration with
$\epsilon=5\times10^{-3}$
and the 20\% confidence gate.
The auxiliary LM-head
readouts incur approximately
0.27 loop-equivalents per pass
on CODI-style LOTUS-1B
and 0.14 on the 3B models.
This overhead is excluded
from the reported loop count.

\paragraph{Huginn KL Exit.}
Huginn~\citep{geiping2025scaling}
supports adaptive recurrent
computation using the KL
divergence between successive
output distributions.
We apply its early-exit
criterion to recurrent
positions:
\begin{equation}
D_{\mathrm{KL}}
\left(
p_i^{(t-1)}
\Vert p_i^{(t)}
\right)
<\epsilon.
\end{equation}
For LOTUS, we obtain the
distributions through
latent-level LM-head readouts
and use the published threshold
$\epsilon=5\times10^{-4}$.
The same distributional-stability
criterion is evaluated in
recurrent prompt prefilling
where applicable.
As with SureLock, auxiliary
LM-head readouts are excluded
from the reported recurrent
computation metric.

\paragraph{Inactive Default Exits.}
We additionally evaluate
the state-convergence exit
of~\citet{geiping2026efficient},
which freezes a position when
its normalized update falls
below $\epsilon=0.03$,
as well as GCEE under its
published defaults and
validation-free configuration.
These rules do not trigger
freezing on the evaluated
LOTUS configurations, yielding
the same predictions and
recurrent computation as
dense inference.
We therefore report the
calibrated GCEE variant
in the main comparison.

\section{Additional Analyses}
\label{app:analysis}

\subsection{Predictor Fit on Held-Out Samples}
\label{app:fit}

\begin{table}[h]
\centering
\setlength{\tabcolsep}{3pt}
\resizebox{\columnwidth}{!}{%
\begin{tabular}{@{}lcccccc@{}}
\toprule
& \multicolumn{2}{c}{\textbf{LOTUS-1B (CODI)}} & \multicolumn{2}{c}{\textbf{LOTUS-3B (CODI)}} & \multicolumn{2}{c}{\textbf{LOTUS-3B (Coconut)}} \\
\cmidrule(l{2pt}r{2pt}){2-3} \cmidrule(l{2pt}r{2pt}){4-5} \cmidrule(l{2pt}r{2pt}){6-7}
\textbf{Signal} & Reus. & Impact & Reus. & Impact & Reus. & Impact \\
\midrule
Last-step K/V change & 0.77 & 0.62 & 0.69 & 0.60 & 0.82 & 0.64 \\
Last-step hidden-state change & 0.77 & 0.58 & 0.75 & 0.55 & 0.84 & 0.60 \\
\midrule
SwiftLoop (reusability) & 0.97 & 0.65 & 0.91 & 0.65 & 0.94 & 0.70 \\
SwiftLoop (impact, IG) & 0.82 & 0.76 & 0.71 & 0.70 & 0.83 & 0.76 \\
SwiftLoop (impact, exact) & 0.82 & 0.76 & 0.75 & 0.78 & 0.82 & 0.80 \\
\bottomrule
\end{tabular}}
\caption{\textbf{Learned versus retrospective freezing signals.}
Within-pass AUC for identifying reusable positions
($E_i^{\mathrm{KV}}(t)<0.45$) and positions with negligible
functional impact ($I_i^{(t)}<1\%$), averaged over
$t=1,\ldots,5$. Learned predictors are evaluated
out of fold, with functional impact assessed
against exact counterfactual labels.}
\label{tab:retro}
\end{table}

\begin{table}[h]
\centering
\setlength{\tabcolsep}{3pt}
\resizebox{\columnwidth}{!}{%
\begin{tabular}{@{}lcccccc@{}}
\toprule
& \multicolumn{2}{c}{\textbf{Reusability}} & \multicolumn{2}{c}{\textbf{Impact (IG)}} & \multicolumn{2}{c}{\textbf{Impact (exact)}} \\
\cmidrule(l{2pt}r{2pt}){2-3} \cmidrule(l{2pt}r{2pt}){4-5} \cmidrule(l{2pt}r{2pt}){6-7}
\textbf{Model} & $\rho$ & AUC & $\rho$ & AUC & $\rho$ & AUC \\
\midrule
LOTUS-1B (CODI) & 0.98 & 0.97 & 0.77 & 0.75 & 0.69 & 0.76 \\
LOTUS-3B (CODI) & 0.94 & 0.91 & 0.63 & 0.68 & 0.77 & 0.75 \\
LOTUS-3B (Coconut) & 0.89 & 0.94 & 0.65 & 0.75 & 0.70 & 0.80 \\
\bottomrule
\end{tabular}}
\caption{\textbf{Out-of-fold predictor evaluation.}
Results on 500 GSM8K validation problems.
$\rho$ denotes Spearman correlation with
regression targets across all position--pass pairs.
AUC measures within-pass identification of
reusable positions ($E_i^{\mathrm{KV}}(t)<0.45$)
or negligible exact functional impact
($I_i^{(t)}<1\%$), averaged over $t=0,\ldots,5$.}
\label{tab:fit}
\end{table}

We evaluate SwiftLoop's predictors using
five-fold out-of-fold predictions on
500 GSM8K validation problems.
We report Spearman correlation with
regression targets and within-pass AUC
for identifying reusable positions
($E_i^{\mathrm{KV}}(t)<0.45$) or
positions with negligible functional
impact ($I_i^{(t)}<1\%$).
AUC is averaged across passes, with
$t=0$ excluded when comparing against
last-step change signals.
Table~\ref{tab:retro} demonstrates the
advantages of hindsight-guided learning
over retrospective state-change signals.
Learned reusability predictors achieve
AUCs of 0.91--0.97, compared with
0.69--0.84 for local K/V and hidden-state
changes.
Likewise, learned impact predictors
achieve AUCs of 0.70--0.80, compared
with 0.55--0.64.
These results demonstrate the value of
learning future-oriented freezing
criteria from current hidden states.
Table~\ref{tab:fit} further demonstrates
the effectiveness of learned freezing
signals across LOTUS models.
Reusability predictors achieve pooled
Spearman correlations of 0.89--0.98
and within-pass AUCs of 0.91--0.97.
For output-level supervision,
IG-based joint-freezing attribution
provides an informative training
signal, achieving AUCs of 0.68--0.75
against exact single-position impact
labels without requiring individual
counterfactual rollouts.

\begin{table*}[t]
\centering
\def\tabdelta#1{{\scriptsize(#1)}}
\resizebox{0.9\textwidth}{!}{%
\begin{tabular}{@{}lccr@{\,}lrr@{\,}lrr@{\,}lrr@{\,}lr@{}}
\toprule
& & & \multicolumn{3}{c}{\textbf{GSM8K}} & \multicolumn{3}{c}{\textbf{SVAMP}} & \multicolumn{3}{c}{\textbf{MultiArith}} & \multicolumn{3}{c}{\textbf{GSM-Hard}} \\
\cmidrule(l{2pt}){4-6} \cmidrule(l{2pt}){7-9} \cmidrule(l{2pt}){10-12} \cmidrule(l{2pt}){13-15}
\textbf{Model} & $K$ & $\rho_{K=4}$ & \multicolumn{2}{c}{Acc. (\%)} & Loops & \multicolumn{2}{c}{Acc. (\%)} & Loops & \multicolumn{2}{c}{Acc. (\%)} & Loops & \multicolumn{2}{c}{Acc. (\%)} & Loops \\
\midrule
LOTUS-1B (CODI) & 4 & -- & 58.83 & \tabdelta{$-$0.5} & 1.80 & 60.10 & \tabdelta{$-$1.0} & 0.79 & 96.67 & \tabdelta{0.0} & 0.98 & 13.42 & \tabdelta{$-$0.2} & 1.52 \\
 & 3 & 0.99 & 58.53 & \tabdelta{$-$0.8} & 1.56 & 59.10 & \tabdelta{$-$2.0} & 0.70 & 96.11 & \tabdelta{$-$0.6} & 0.87 & 13.34 & \tabdelta{$-$0.2} & 1.36 \\
 & 2 & 0.97 & 58.98 & \tabdelta{$-$0.3} & 1.88 & 60.30 & \tabdelta{$-$0.8} & 0.83 & 96.67 & \tabdelta{0.0} & 1.02 & 13.34 & \tabdelta{$-$0.2} & 1.58 \\
\midrule
LOTUS-3B (CODI) & 4 & -- & 71.11 & \tabdelta{+0.4} & 5.05 & 75.80 & \tabdelta{+0.2} & 3.54 & 100.00 & \tabdelta{+0.6} & 3.97 & 16.38 & \tabdelta{0.0} & 5.20 \\
 & 3 & 0.76 & 70.74 & \tabdelta{0.0} & 5.37 & 75.50 & \tabdelta{$-$0.1} & 3.75 & 100.00 & \tabdelta{+0.6} & 4.39 & 16.30 & \tabdelta{$-$0.1} & 5.56 \\
 & 2 & 0.71 & 70.89 & \tabdelta{+0.2} & 5.08 & 75.60 & \tabdelta{0.0} & 3.64 & 100.00 & \tabdelta{+0.6} & 4.07 & 16.45 & \tabdelta{+0.1} & 5.20 \\
\midrule
LOTUS-3B (Coconut) & 4 & -- & 70.20 & \tabdelta{0.0} & 5.04 & 73.70 & \tabdelta{$-$0.2} & 3.63 & 99.44 & \tabdelta{0.0} & 3.05 & 16.53 & \tabdelta{+0.1} & 4.52 \\
 & 3 & 0.85 & 70.43 & \tabdelta{+0.2} & 5.20 & 73.70 & \tabdelta{$-$0.2} & 3.93 & 98.89 & \tabdelta{$-$0.6} & 3.91 & 16.53 & \tabdelta{+0.1} & 5.29 \\
 & 2 & 0.76 & 70.20 & \tabdelta{0.0} & 4.87 & 73.30 & \tabdelta{$-$0.6} & 3.49 & 98.89 & \tabdelta{$-$0.6} & 3.56 & 16.68 & \tabdelta{+0.2} & 4.94 \\
\bottomrule
\end{tabular}}
\caption{\textbf{Ablation on IG quadrature steps.}
$\rho_{K=4}$ denotes the within-pass Spearman correlation
of attribution labels with those obtained using $K=4$,
evaluated on 500 GSM8K validation problems.
We report test accuracy (with changes relative to
fixed six-loop inference) and average loops per latent position.}
\label{tab:igk}
\end{table*}

\subsection{Ablation on IG Quadrature Steps}
\label{app:igk}

We approximate the integrated-gradient
path integral in Eq.~\eqref{eq:ig_impact}
using $K$ midpoint nodes.
Each node requires one forward--backward
pass per recurrent step, jointly
attributing all positions.
Table~\ref{tab:igk} examines
$K\in\{2,3,4\}$ on LOTUS, with
predictors refitted and freezing
thresholds recalibrated for each setting.
The resulting attribution labels exhibit
consistent rankings across quadrature
resolutions, with within-pass Spearman
correlations of 0.71--0.99 relative
to $K=4$.
The learned policies also maintain
comparable accuracy--computation
trade-offs across benchmarks.

\begin{table}[h]
\centering
\setlength{\tabcolsep}{3pt}
\resizebox{\columnwidth}{!}{%
\begin{tabular}{@{}llcccc@{}}
\toprule
\textbf{Model} & \textbf{Data set} & \textbf{Batch size} & \textbf{Timed} & \textbf{Input tokens} & \textbf{Choices} \\
\midrule
\multirow{3}{*}{LOTUS (all)} & GSM8K & 256 & 256 & 59--60 / 133 & -- \\
 & SVAMP & 256 & 256 & 40 / 65 & -- \\
 & MultiArith & 256 & 180 & 40 / 54 & -- \\
\midrule
\multirow{3}{*}{Ouro-1.4B} & WinoGrande & 16 & 128 & 138 / 163 & 2 \\
 & ARC-C & 4 & 32 & 992 / 1207 & 4 \\
 & HellaSwag & 16 & 64 & 830 / 1053 & 4 \\
\midrule
\multirow{3}{*}{Huginn-3.5B} & WinoGrande & 16 & 64 & 134 / 158 & 2 \\
 & ARC-C & shared context & 32 & 978 / 1191 & 4 \\
 & HellaSwag & shared context & 32 & 805 / 975 & 4 \\
\bottomrule
\end{tabular}}
\caption{\textbf{Latency measurement configurations.}
Batch sizes, numbers of timed test instances,
and input lengths (mean / maximum, in tokens).
Input lengths refer to questions for LOTUS
and few-shot contexts per answer choice
for recurrent prefilling.
Shared context denotes single-context
prefilling with K/V reuse across choices.}
\label{tab:latency-setup}
\end{table}

\begin{table*}[t]
\centering
\setlength{\tabcolsep}{6pt}
\resizebox{0.7\textwidth}{!}{%
\begin{tabular}{@{}llcccc@{}}
\toprule
& & \multicolumn{2}{c}{\textbf{SVAMP}}
& \multicolumn{2}{c}{\textbf{MultiArith}} \\
\cmidrule(l{2pt}r{2pt}){3-4}
\cmidrule(l{2pt}r{2pt}){5-6}
\textbf{Model} & \textbf{Setting}
& Recurrent & End-to-end
& Recurrent & End-to-end \\
\midrule
\multirow{4}{*}{LOTUS-1B (CODI)}
& Fixed (dense) & -- & 25.1 & -- & 24.5 \\
& Fixed (sparse) & 19.3 & 25.7 & 19.3 & 25.1 \\
& SwiftLoop (reusability) & 6.8 & 13.0 & 6.9 & 12.7 \\
& SwiftLoop (impact, IG) & 3.5 & 11.1 & 4.1 & 9.9 \\
\midrule
\multirow{4}{*}{LOTUS-3B (CODI)}
& Fixed (dense) & -- & 62.8 & -- & 61.6 \\
& Fixed (sparse) & 49.1 & 64.4 & 49.1 & 63.2 \\
& SwiftLoop (reusability) & 37.2 & 52.5 & 43.6 & 57.7 \\
& SwiftLoop (impact, IG) & 30.3 & 45.6 & 33.7 & 47.8 \\
\midrule
\multirow{4}{*}{LOTUS-3B (Coconut)}
& Fixed (dense) & -- & 45.6 & -- & 44.6 \\
& Fixed (sparse) & 36.3 & 47.2 & 36.3 & 46.1 \\
& SwiftLoop (reusability) & 31.4 & 42.3 & 31.4 & 41.3 \\
& SwiftLoop (impact, IG) & 23.4 & 34.3 & 19.9 & 29.7 \\
\bottomrule
\end{tabular}}
\caption{\textbf{Latency breakdown for implicit reasoning}
(ms per problem; batch size 256).
Recurrent denotes the six-loop computation,
including freezing decisions;
end-to-end includes question prefilling,
latent initialization, and answer decoding.
Dense inference is timed end to end.}
\label{tab:latency-abs}
\end{table*}

\begin{table}[h]
\centering
\setlength{\tabcolsep}{4pt}
\begin{tabular}{@{}llccc@{}}
\toprule
\textbf{Model} & \textbf{Task} & Dense & Reus. & IG \\
\midrule
\multirow{3}{*}{Ouro-1.4B} & WinoGrande & 27.8 & 24.9 & 19.4 \\
 & ARC-C & 402.1 & 257.1 & 247.7 \\
 & HellaSwag & 322.4 & 212.1 & 173.3 \\
\midrule
\multirow{3}{*}{Huginn-3.5B} & WinoGrande & 204.0 & 111.8 & 138.0 \\
 & ARC-C & 922.6 & 630.9 & 732.1 \\
 & HellaSwag & 811.9 & 552.1 & 646.6 \\
\bottomrule
\end{tabular}
\caption{\textbf{Absolute latency of recurrent prompt prefilling}
(ms per document).
Measurements cover scoring all answer choices
at the batch sizes in Table~\ref{tab:latency-setup}.
Shared contexts are prefixed once per document
on Huginn.}
\label{tab:latency-prefill}
\end{table}

\subsection{Computation Accounting}
\label{app:accounting}

\noindent\textbf{Recurrent Steps.}
For LOTUS, Loops denotes the average
number of recurrent updates per position
across 150 latent slots and the start token,
excluding the initialization pass.
For recurrent prompt prefilling, Steps
denotes the average recurrent-block
applications per context token.

\noindent\textbf{FLOPs.}
Figure~\ref{fig:flop} reports analytical
FLOPs per problem, normalized to
fixed six-loop inference, with two
FLOPs per multiply--accumulate.
The accounting includes question
prefilling, latent initialization,
recurrent refinement, and eight answer
decoding steps, including the LM head.
Recurrent FLOPs account for projections
and MLPs at active positions, as well
as attention to all visible K/V,
including frozen positions.
Policy-specific overhead is excluded
from the analytical FLOP estimates.
Recurrent refinement accounts for
approximately 80\% of dense inference
FLOPs across the LOTUS models.

\noindent\textbf{Wall-Clock Latency.}
Table~\ref{tab:latency-all} reports
measured end-to-end inference speedups,
including predictor evaluation and
freezing decisions.
Further measurement details are provided
in Appendix~\ref{app:latency}.

\subsection{Latency Measurement}
\label{app:latency}

\paragraph{Setup.}
All experiments run on a single NVIDIA
A100 80GB PCIe GPU with PyTorch 2.7.1,
CUDA 12.6, and bfloat16 computation.
We compare each model's native dense
Hugging Face forward against a sparse
executor that updates only active positions
while preserving frozen K/V for attention.
Without freezing, the sparse executor
achieves 0.96--1.00$\times$ dense
inference speed, with predictor evaluation
adding at most 3\% overhead.

\paragraph{Timing Protocol.}
For LOTUS, timing covers the full
inference pipeline, from question
tokenization through prefill,
latent initialization, recurrent
refinement, and answer decoding.
For Ouro and Huginn, timing covers
end-to-end answer-choice scoring,
excluding tokenization.
Huginn reuses the prefilling cache
of shared contexts across answer choices.
We synchronize CUDA at timing boundaries
and report median wall-clock latency
over three rounds following warm-up.
Speedups are computed from median
latencies for LOTUS and median per-round
ratios for prompt prefilling.
Table~\ref{tab:latency-setup}
summarizes the timing configurations.

\paragraph{Latency Breakdown.}
Table~\ref{tab:latency-abs} reports
recurrent and end-to-end latency
for implicit reasoning.
On LOTUS-1B, SwiftLoop (impact, IG)
accelerates recurrent refinement
by $4.7$--$5.5\times$ on SVAMP
and MultiArith, yielding
$2.26$--$2.48\times$ end-to-end speedups.
Table~\ref{tab:latency-prefill}
further reports absolute latencies
for recurrent prompt prefilling.
On Huginn, selective refinement achieves
1.46--1.47$\times$ acceleration on
ARC-Challenge and HellaSwag even with
single-context prefilling, demonstrating
effective sparse execution on long contexts.

\section{Additional Baselines Comparison}

\subsection{Unified Baseline Calibration}
Table~\ref{tab:main-noloss} follows
the published calibration procedures
or default settings of adaptive baselines,
except H2O and LazyLLM, which use
SwiftLoop's validation criterion.
For a controlled comparison, we
recalibrate FlashLoop, CALM, GCEE,
SureLock, and the Huginn KL exit
on the same validation samples.
For each model, we select the setting
that minimizes recurrent computation
while preserving fixed-loop validation
accuracy ($\eta\le0.5$), or the
highest-accuracy setting if none qualifies.
We search the original CALM and GCEE
parameter grids, FlashLoop reconstruction
bounds of 5--80\%, and twelve thresholds
for SureLock and the KL exit.
Selected settings remain fixed
across test benchmarks.

\noindent\textbf{Selected settings.}
For LOTUS-1B (CODI), LOTUS-3B (CODI),
and LOTUS-3B (Coconut), respectively:
FlashLoop bounds are $(20,5,5)\%$;
CALM thresholds are
$(0.937,0.969,0.974)$;
GCEE parameters are
$(0.726,0.5)$, $(0.566,0)$,
and $(0.250,0)$;
SureLock thresholds are
$(10^{-2},2\times10^{-2},2\times10^{-3})$;
and KL-exit thresholds are
$(10^{-2},10^{-3},5\times10^{-4})$.

\noindent\textbf{Results.}
Table~\ref{tab:unical} demonstrates that
SwiftLoop maintains its advantage
under unified baseline calibration.
Across all 12 model--benchmark settings,
SwiftLoop achieves greater recurrent
computation savings than the
recalibrated adaptive baselines,
while preserving competitive accuracy.
These results further validate the
effectiveness of hindsight-guided
selective refinement.

\begin{table*}[t]
\centering
\setlength{\tabcolsep}{4pt}
\def\tabdelta#1{{\scriptsize(#1)}}
\resizebox{0.85\textwidth}{!}{%
\begin{tabular}{@{}llr@{\,}lrr@{\,}lrr@{\,}lrr@{\,}lr@{}}
\toprule
& & \multicolumn{3}{c}{\textbf{GSM8K}} & \multicolumn{3}{c}{\textbf{SVAMP}} & \multicolumn{3}{c}{\textbf{MultiArith}} & \multicolumn{3}{c}{\textbf{GSM-Hard}} \\
\cmidrule(l{2pt}){3-5} \cmidrule(l{2pt}){6-8} \cmidrule(l{2pt}){9-11} \cmidrule(l{2pt}){12-14}
\textbf{Model} & \textbf{Method} & \multicolumn{2}{c}{Acc. (\%)} & Loops & \multicolumn{2}{c}{Acc. (\%)} & Loops & \multicolumn{2}{c}{Acc. (\%)} & Loops & \multicolumn{2}{c}{Acc. (\%)} & Loops \\
\midrule
LOTUS-1B (CODI) & Fixed (6 loops) & 59.29 &  & 6.00 & 61.10 &  & 6.00 & 96.67 &  & 6.00 & 13.57 &  & 6.00 \\
 & FlashLoop & 59.36 & \tabdelta{+0.1} & 4.77 & 61.30 & \tabdelta{+0.2} & 4.77 & 97.22 & \tabdelta{+0.6} & 4.77 & 13.57 & \tabdelta{0.0} & 4.77 \\
 & CALM & 59.51 & \tabdelta{+0.2} & 3.25 & 60.60 & \tabdelta{$-$0.5} & 2.68 & 97.22 & \tabdelta{+0.6} & 2.91 & 13.65 & \tabdelta{+0.1} & 3.33 \\
 & GCEE & 58.76 & \tabdelta{$-$0.5} & 3.74 & 60.40 & \tabdelta{$-$0.7} & 3.27 & 97.22 & \tabdelta{+0.6} & 3.32 & 13.27 & \tabdelta{$-$0.3} & 3.71 \\
 & SureLock & 59.06 & \tabdelta{$-$0.2} & 5.86 & 61.20 & \tabdelta{+0.1} & 5.90 & 96.67 & \tabdelta{0.0} & 5.86 & 13.50 & \tabdelta{$-$0.1} & 5.90 \\
 & Huginn KL exit$^{\ddagger}$ & 59.29 & \tabdelta{0.0} & 5.84 & 61.00 & \tabdelta{$-$0.1} & 5.89 & 96.67 & \tabdelta{0.0} & 5.87 & 13.57 & \tabdelta{0.0} & 5.88 \\
\midrule
LOTUS-3B (CODI) & Fixed (6 loops) & 70.74 &  & 6.00 & 75.60 &  & 6.00 & 99.44 &  & 6.00 & 16.38 &  & 6.00 \\
 & FlashLoop & 70.51 & \tabdelta{$-$0.2} & 5.95 & 75.70 & \tabdelta{+0.1} & 5.95 & 99.44 & \tabdelta{0.0} & 5.95 & 16.45 & \tabdelta{+0.1} & 5.95 \\
 & CALM & 70.58 & \tabdelta{$-$0.2} & 5.90 & 75.50 & \tabdelta{$-$0.1} & 5.58 & 99.44 & \tabdelta{0.0} & 5.78 & 16.38 & \tabdelta{0.0} & 5.93 \\
 & GCEE & 70.66 & \tabdelta{$-$0.1} & 5.79 & 75.90 & \tabdelta{+0.3} & 5.45 & 99.44 & \tabdelta{0.0} & 5.66 & 16.45 & \tabdelta{+0.1} & 5.80 \\
 & SureLock & 70.74 & \tabdelta{0.0} & 5.96 & 75.50 & \tabdelta{$-$0.1} & 5.96 & 99.44 & \tabdelta{0.0} & 5.94 & 16.38 & \tabdelta{0.0} & 5.97 \\
 & Huginn KL exit & 70.66 & \tabdelta{$-$0.1} & 5.99 & 75.60 & \tabdelta{0.0} & 6.00 & 99.44 & \tabdelta{0.0} & 5.99 & 16.45 & \tabdelta{+0.1} & 5.99 \\
\midrule
LOTUS-3B (Coconut) & Fixed (6 loops) & 70.20 &  & 6.00 & 73.90 &  & 6.00 & 99.44 &  & 6.00 & 16.45 &  & 6.00 \\
 & FlashLoop & 70.20 & \tabdelta{0.0} & 6.00 & 73.90 & \tabdelta{0.0} & 6.00 & 99.44 & \tabdelta{0.0} & 6.00 & 16.45 & \tabdelta{0.0} & 6.00 \\
 & CALM$^{\ddagger}$ & 70.20 & \tabdelta{0.0} & 5.85 & 73.90 & \tabdelta{0.0} & 5.44 & 99.44 & \tabdelta{0.0} & 5.69 & 16.30 & \tabdelta{$-$0.2} & 5.87 \\
 & GCEE & 70.20 & \tabdelta{0.0} & 5.95 & 73.90 & \tabdelta{0.0} & 5.78 & 99.44 & \tabdelta{0.0} & 5.89 & 16.45 & \tabdelta{0.0} & 5.96 \\
 & SureLock & 69.07 & \tabdelta{$-$1.1} & 5.96 & 73.30 & \tabdelta{$-$0.6} & 5.97 & 99.44 & \tabdelta{0.0} & 5.96 & 16.22 & \tabdelta{$-$0.2} & 5.96 \\
 & Huginn KL exit & 69.14 & \tabdelta{$-$1.1} & 5.97 & 73.60 & \tabdelta{$-$0.3} & 5.97 & 99.44 & \tabdelta{0.0} & 5.96 & 16.15 & \tabdelta{$-$0.3} & 5.97 \\
\bottomrule
\end{tabular}}
\caption{\textbf{Adaptive baselines under unified calibration.} Test accuracy (change relative to fixed six-loop inference) and average loops per latent position.
Each method is calibrated on validation set
to minimize recurrent computation with at most a 0.5 percentage point accuracy drop.
$^{\ddagger}$ denotes the highest-accuracy
setting when no candidate satisfies the
constraint.
SwiftLoop, H2O, and LazyLLM results under
the same criterion appear in
Table~\ref{tab:main-noloss}.}
\label{tab:unical}
\end{table*}

\begin{table*}[t]
\centering
\resizebox{0.85\textwidth}{!}{%
\begin{tabular}{@{}llrrrrrrrr@{}}
\toprule
& & \multicolumn{2}{c}{\textbf{GSM8K}} & \multicolumn{2}{c}{\textbf{SVAMP}} & \multicolumn{2}{c}{\textbf{MultiArith}} & \multicolumn{2}{c}{\textbf{GSM-Hard}} \\
\cmidrule(l{2pt}){3-4} \cmidrule(l{2pt}){5-6} \cmidrule(l{2pt}){7-8} \cmidrule(l{2pt}){9-10}
\textbf{Model} & \textbf{Method} & Acc. (\%) & Loops & Acc. (\%) & Loops & Acc. (\%) & Loops & Acc. (\%) & Loops \\
\midrule
LOTUS-1B (CODI) & Fixed (6 loops) & 59.29 & 6.00 & 61.10 & 6.00 & 96.67 & 6.00 & 13.57 & 6.00 \\
 & Random (IG budget) & 46.40\rlap{*} & 1.79 & 55.20\rlap{*} & 0.79 & 83.89\rlap{*} & 0.98 & 9.40\rlap{*} & 1.52 \\
 & Random (exact budget) & 45.41\rlap{*} & 1.54 & 56.10\rlap{*} & 0.68 & 76.67\rlap{*} & 0.79 & 9.78\rlap{*} & 1.40 \\
\midrule
LOTUS-3B (CODI) & Fixed (6 loops) & 70.74 & 6.00 & 75.60 & 6.00 & 99.44 & 6.00 & 16.38 & 6.00 \\
 & Random (IG budget) & 69.14\rlap{*} & 5.07 & 75.00 & 3.52 & 100.00 & 3.98 & 16.07 & 5.21 \\
 & Random (exact budget) & 68.76\rlap{*} & 5.03 & 74.10\rlap{*} & 3.46 & 100.00 & 3.38 & 16.38 & 5.45 \\
\midrule
LOTUS-3B (Coconut) & Fixed (6 loops) & 70.20 & 6.00 & 73.90 & 6.00 & 99.44 & 6.00 & 16.45 & 6.00 \\
 & Random (IG budget) & 67.55\rlap{*} & 5.04 & 69.90\rlap{*} & 3.63 & 97.22 & 3.05 & 14.56\rlap{*} & 4.52 \\
 & Random (exact budget) & 67.55\rlap{*} & 5.21 & 69.90\rlap{*} & 3.99 & 98.89 & 3.85 & 16.15 & 5.47 \\
\bottomrule
\end{tabular}}
\caption{\textbf{Random freezing at matched computation budgets.} Test accuracy and average loops per latent position.
Random baselines match the per-pass freezing counts of SwiftLoop,
while selecting positions uniformly at random.}
\label{tab:random}
\end{table*}

\subsection{Comparison with Random Freezing}
\label{app:random}
We evaluate the effectiveness of
SwiftLoop's adaptive freezing decisions
against budget-matched random freezing.
For each variant,
we compare against a budget-matched
random baseline that selects active
positions uniformly at random.
Table~\ref{tab:random} shows that
SwiftLoop improves accuracy over
random freezing by 3.6--20.0
percentage points on LOTUS-1B
across reasoning benchmarks.
Consistent advantages are also
observed on both LOTUS-3B configurations.
These results highlight the effectiveness
of learned freezing decisions in
allocating recurrent computation
across samples and positions.

\end{document}